%% file: main.tex
\documentclass{article}

\PassOptionsToPackage{numbers,sort&compress}{natbib}
\usepackage[eandd,preprint]{neurips_2026}

\usepackage[utf8]{inputenc}
\usepackage[T1]{fontenc}
\usepackage{hyperref}
\usepackage{url}
\usepackage{booktabs}
\usepackage{threeparttable}
\usepackage{amsfonts}
\usepackage{amssymb}
\usepackage{amsmath}
\usepackage{nicefrac}
\usepackage{microtype}
\usepackage[table]{xcolor}
\usepackage{graphicx}
\usepackage{tikz}
\usepackage{float}
\usepackage{placeins}
\usepackage{fontawesome5}
\usepackage[most]{tcolorbox}
\graphicspath{{figures/}}

\definecolor{BVBest}{HTML}{F3E1F4}
\definecolor{BVSecond}{HTML}{DFF6FB}
\definecolor{BVClosedBand}{HTML}{F1F6EC}
\definecolor{BVGeneralBand}{HTML}{FCFAED}
\definecolor{BVMedicalBand}{HTML}{FFF1F3}
\definecolor{BVNativeBand}{HTML}{EFFAF7}
\definecolor{BVBandText}{HTML}{6F7370}
\definecolor{BVIconFull}{HTML}{238B2E}
\definecolor{BVIconPartial}{HTML}{C7A36B}
\definecolor{BVIconNone}{HTML}{AEB9AF}
\definecolor{BVHuatuoBadge}{HTML}{2F9E73}
\definecolor{BVRadFMBadge}{HTML}{5470C6}

\newcommand{\bvrankbest}[1]{\textbf{#1}}
\newcommand{\bvranksecond}[1]{\underline{#1}}
\newcommand{\bvoverallbest}[1]{\cellcolor{BVBest}\textbf{#1}}
\newcommand{\bvoverallsecond}[1]{\cellcolor{BVSecond}\underline{#1}}
\newcommand{\bvgroup}[3]{\multicolumn{#1}{c}{\cellcolor{#2}\color{BVBandText}\textbf{\textit{#3}}} \\}
\newcommand{\bvstatusicon}[2]{%
  \tikz[baseline=-0.35ex]{%
    \node[circle, fill=#1, text=white, minimum size=0.82em,
      inner sep=0pt, outer sep=0pt, font=\tiny\bfseries] {\scalebox{0.78}{#2}};%
  }%
}
\newcommand{\bvfull}{\bvstatusicon{BVIconFull}{\ensuremath{\checkmark}}}
\newcommand{\bvpartial}{\textcolor{BVIconPartial}{\raisebox{0.06ex}{\scalebox{0.78}{$\circ$}}}}
\newcommand{\bvnone}{\bvstatusicon{BVIconNone}{\ensuremath{\times}}}
\newcommand{\tabcheck}{\raisebox{0.08ex}{\scalebox{0.74}{$\checkmark$}}}

\newcommand{\modelicon}[1]{\raisebox{-0.16ex}{\includegraphics[height=0.95em]{#1}}\hspace{0.25em}}

\newcommand{\modelbadge}[2]{%
  \raisebox{0.05ex}{%
    \tikz[baseline=(badge.base)]{%
      \node[
        rounded corners=0.18em,
        fill=#1,
        text=white,
        inner xsep=0.22em,
        inner ysep=0.07em,
        font=\scriptsize\bfseries
      ] (badge) {#2};%
    }%
  }%
  \hspace{0.25em}%
}
\newcommand{\qwenicon}{\modelicon{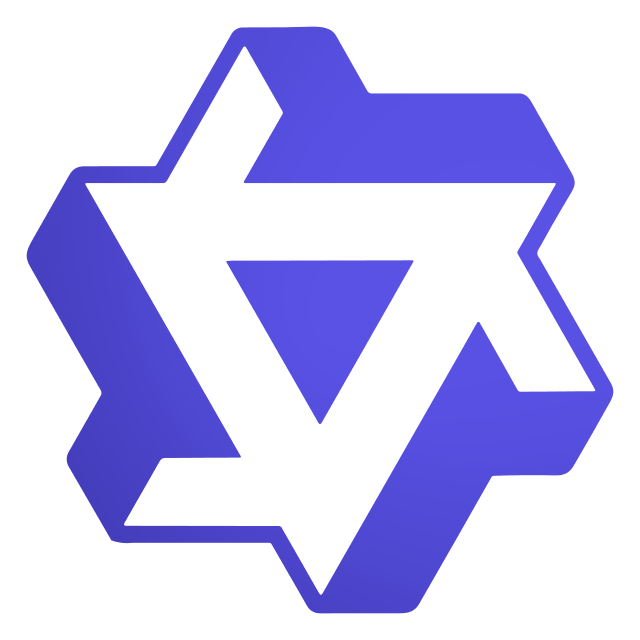}}
\newcommand{\internvlicon}{\modelicon{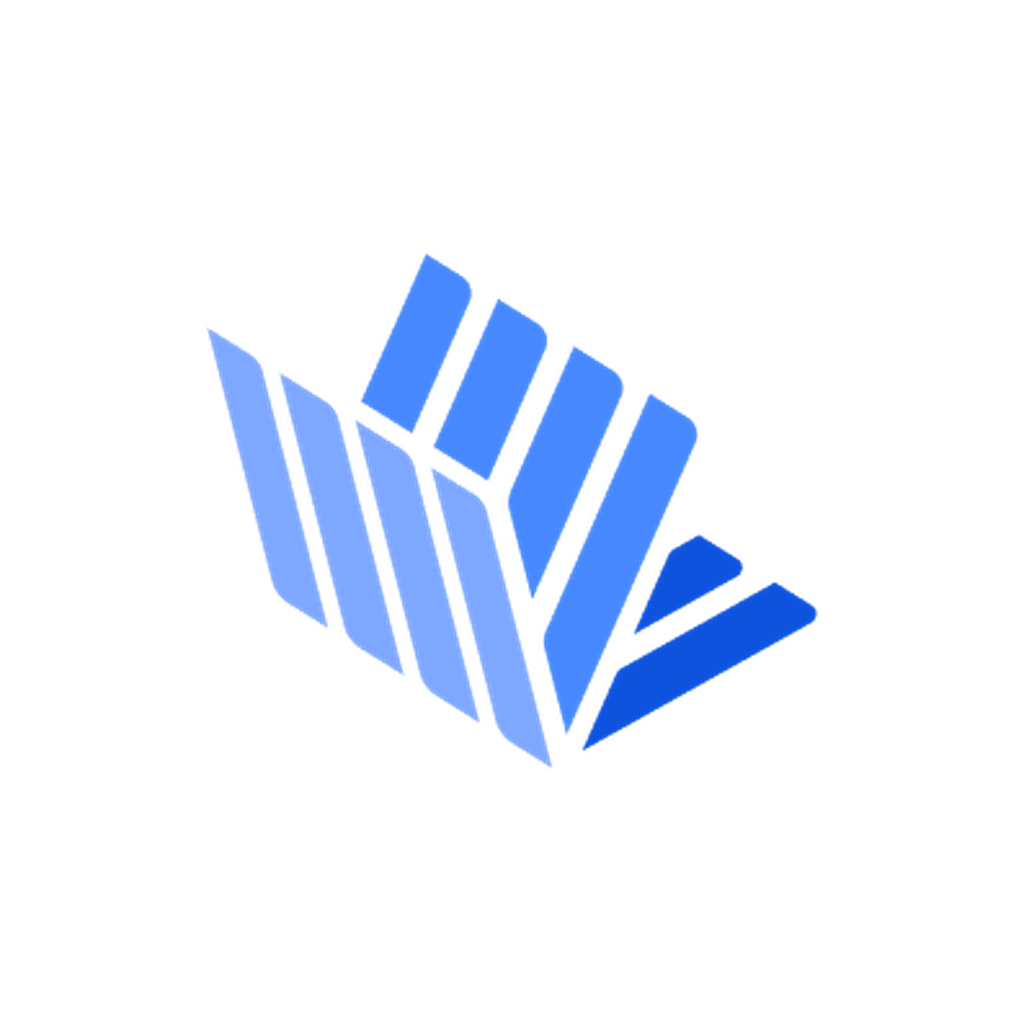}}
\newcommand{\deepseekicon}{\modelicon{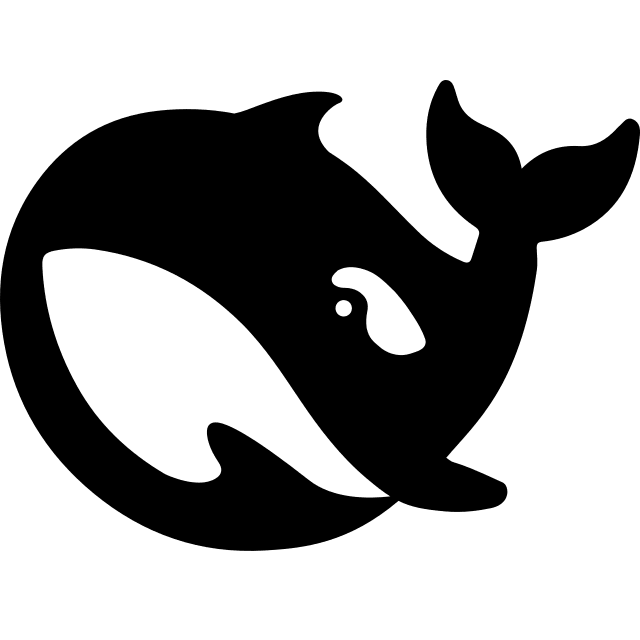}}
\newcommand{\huatuoicon}{\modelbadge{BVHuatuoBadge}{H}}
\newcommand{\lingshuicon}{\modelicon{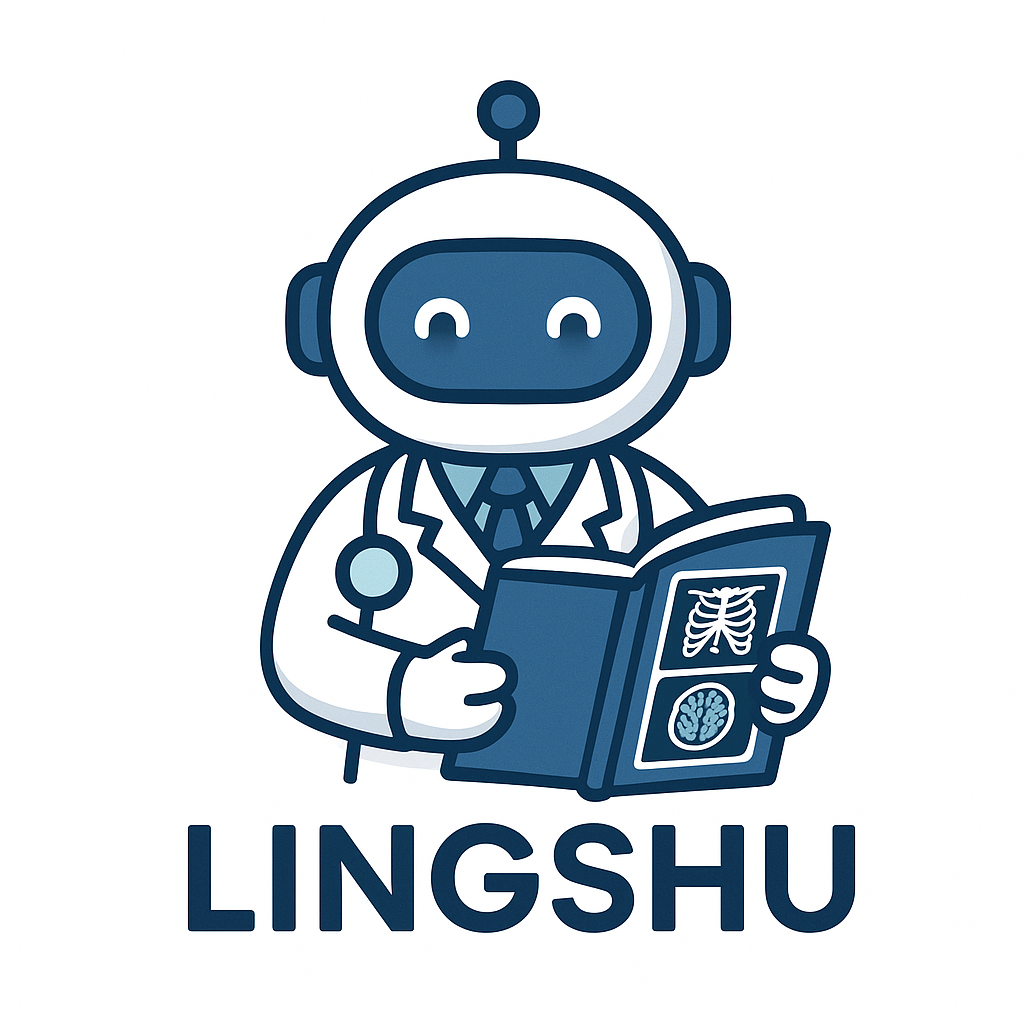}}
\newcommand{\gemmaicon}{\modelicon{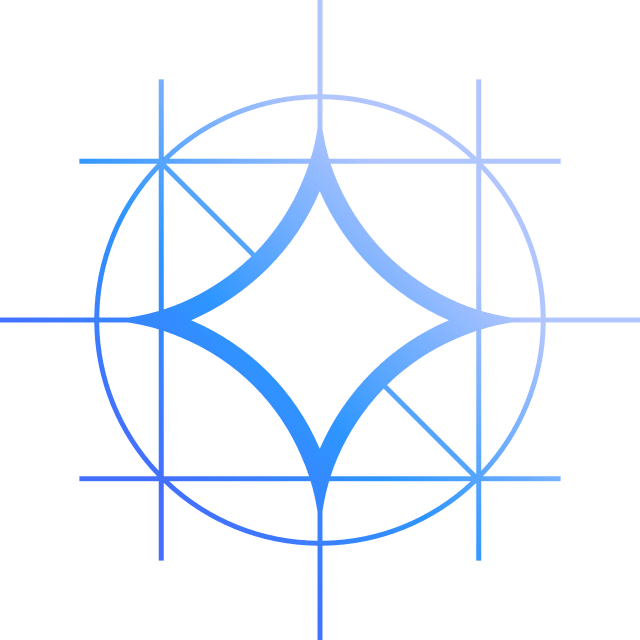}}
\newcommand{\llavaicon}{\modelicon{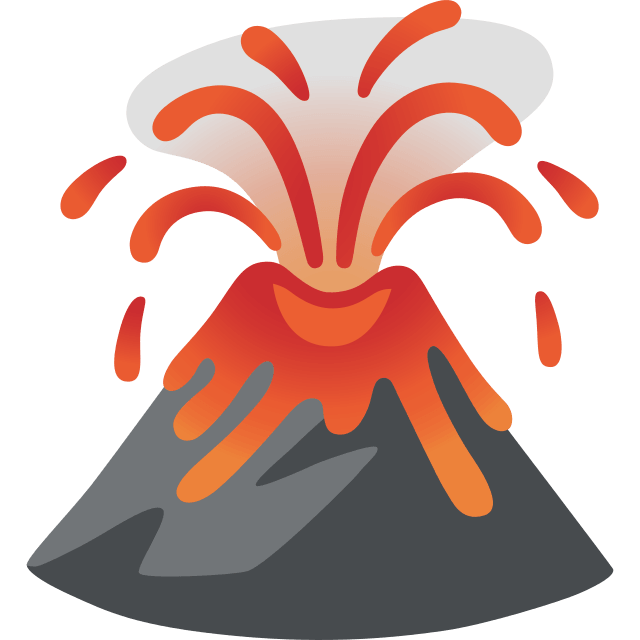}}
\newcommand{\openaiicon}{\modelicon{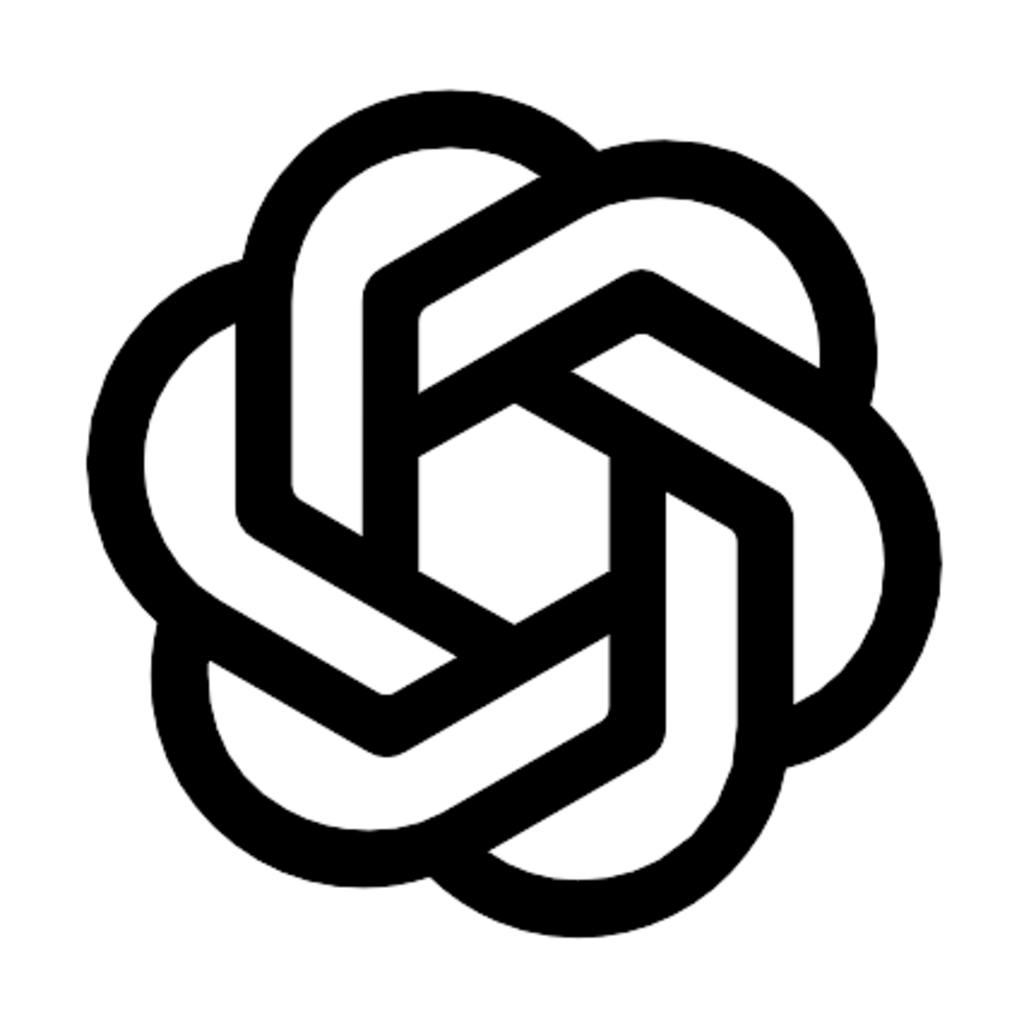}}
\newcommand{\anthropicicon}{\modelicon{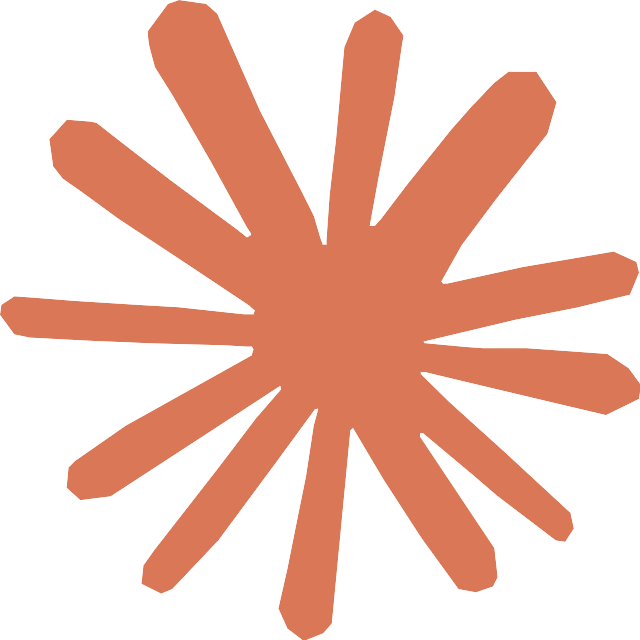}}
\newcommand{\geminiicon}{\modelicon{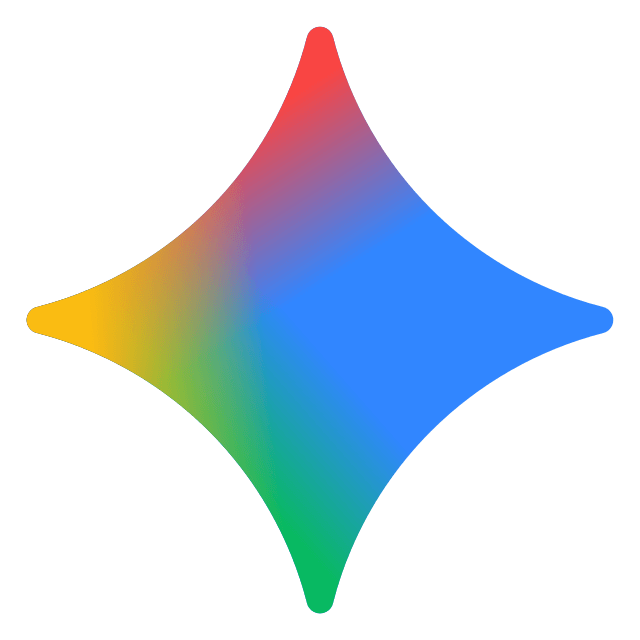}}
\newcommand{\baaiicon}{\modelicon{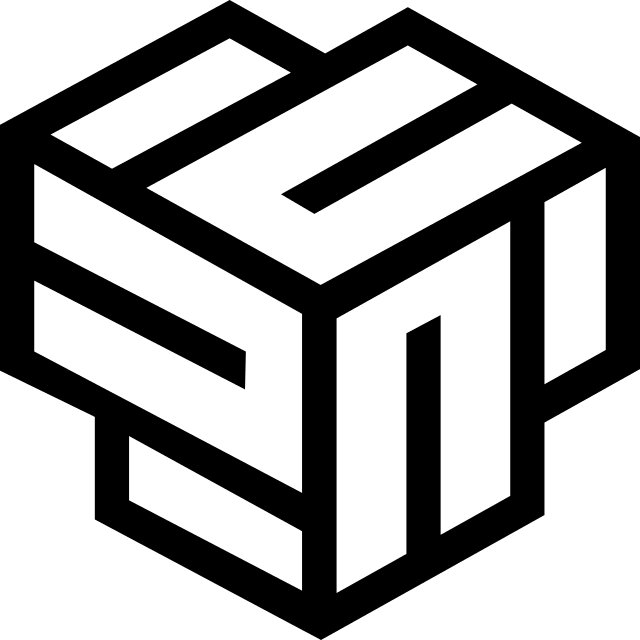}}
\newcommand{\radfmicon}{\modelbadge{BVRadFMBadge}{R}}

\title{BrainTRACE: Tracing Longitudinal, Multimodal, and Volumetric
Evidence in Brain MRI Clinical Reasoning}

\author{%
  Qizhen Lan$^{1,*}$ \quad Mengchen Fan$^{2,*}$ \quad Hang Zhang$^{3,*}$ \quad Jingwei Duan$^{4}$ \\
  \textbf{Moule Lin$^{5}$ \quad Jialin Chen$^{6}$ \quad Baocheng Geng$^{2}$ \quad Xiaoqian Jiang$^{1,\dagger}$} \\[0.5em]
  {\small $^{1}$University of Texas Health Science Center at Houston} \\
  {\small $^{2}$University of Alabama at Birmingham \quad $^{3}$University of Pittsburgh} \\
  {\small $^{4}$The University of Texas MD Anderson Cancer Center} \\
  {\small $^{5}$University of Dublin, Trinity College \quad $^{6}$Yale University} \\
  {\small $^{*}$Equal contribution. \quad $^{\dagger}$Corresponding author: Xiaoqian Jiang.} \\
  {\small Contact: \texttt{qizhen.lan@uth.tmc.edu}}
}

\begin{document}
\maketitle

\begin{abstract}
Brain MRI interpretation is a longitudinal clinical reasoning problem: radiologists compare serial studies, integrate information across MRI sequences, localize findings within volumetric anatomy, and translate this evidence into report-grounded assessments. Existing medical VQA and 3D imaging benchmarks capture important parts of this workflow, but often evaluate brain MRI through isolated images, static volumes, or ungrounded report-style answers, thereby obscuring failures in the evidence chain that support clinical validity. We introduce \textbf{BrainTRACE}, a report-grounded benchmark for evaluating whether vision-language models can trace the evidence structure required for longitudinal brain MRI interpretation. BrainTRACE contains 7,273 scored VQA instances derived from 1,778 longitudinal patients, 7,299 MRI studies, and approximately 29k co-registered 3D MRI sequence volumes. The benchmark is organized by five levels of clinical reasoning, from acquisition recognition to case-level synthesis, and by evidence demands covering longitudinal comparison, report-grounded references, multi-sequence integration, and volumetric spatial evidence. BrainTRACE supports rendered inputs compatible with standard VLM interfaces, a 3D-evidence condition, and a decomposed case-reasoning track that audits six steps in a longitudinal evidence chain. Evaluation of 20 VLM configurations shows that current systems can identify isolated visual cues but rarely compose them into grounded longitudinal interpretations. We release the benchmark specification, evaluation lists, scoring implementation, scoring rubrics, and audit-record format to support reproducible progress in brain MRI VLM evaluation. \footnote{Availability. The dataset release is available at
\url{https://huggingface.co/datasets/lanqz7766/BrainTRACE}. Code and documentation are available at \url{https://github.com/lanqz7766/BrainTRACE}. Machine-readable RAI metadata is shipped as a Croissant file alongside the dataset on HuggingFace.}

\end{abstract}


\input{sections/01_introduction}

\input{sections/02_related_work}

\input{sections/03_benchmark}
\input{sections/04_experiments}

\input{sections/06_conclusion}

\bibliographystyle{plainnat}
\bibliography{references}
\clearpage
\input{sections/A_appendix_gap_table}
\input{sections/B_appendix_template_tables}
\FloatBarrier
\input{sections/C_appendix_ablation}
\FloatBarrier
\input{sections/D_appendix_cases}
\FloatBarrier
\input{sections/05_limitations_ethics}
\end{document}

%% file: sections/01_introduction.tex
\section{Introduction}
\label{sec:intro}
Visual models trained end to end have achieved strong performance in object detection and 3D medical image segmentation~\citep{lan2025acamkd,lan2026recokd,fan2026dcd}. General and medical VLMs extend image analysis to natural-language descriptions, question answering, and visual reasoning~\citep{bai2025qwen25vl,zhu2025internvl3,wu2025radfm,xiao2026vigil,wang2026atlasva}. Disease interpretation on brain MRI, however, requires reasoning beyond an isolated image or sequence. It often depends on jointly evaluating lesion location, sequence-specific appearance, and longitudinal change, including where abnormalities are located, how they appear across different tissue contrasts, and whether they are new, improved, stable, or progressive across serial studies. A trustworthy interpretation should follow an auditable evidence chain rather than merely describe the current disease state. This distinction is especially important for recent VLM studies, where fluent clinical prose can mask failures in true image-grounded reasoning. Studies of large reasoning models show that fluent outputs may reflect memorized patterns rather than input-grounded reasoning~\citep{wang2025reasoningretrievalstudyanswer}. Thus, the key question is not whether a VLM can describe an isolated brain MRI, but whether it can reason from the evidence needed for clinically meaningful follow-up interpretation.

\begin{figure}[t]
    \centering
    \includegraphics[width=\linewidth]{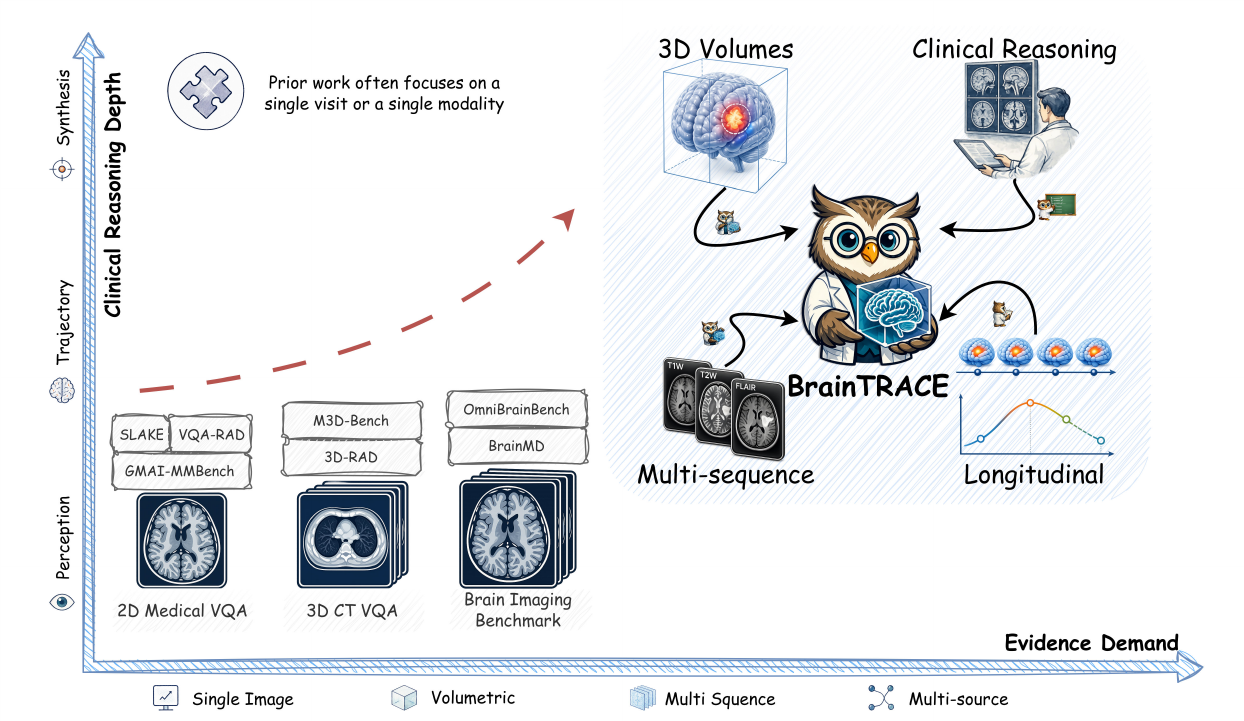}
    \caption{Positioning of BrainTRACE relative to representative medical VLM/VQA benchmarks. BrainTRACE makes longitudinal brain MRI follow-up the central evaluation unit, requiring models to integrate serial, multi-sequence, volumetric, and report-grounded clinical evidence.}
    \label{fig:bench_draw}
\end{figure}

Existing benchmarks cover important parts of this problem, but they have challenges to evaluate the full follow-up reasoning process required in clinical settings. General medical VQA benchmarks
make broad image question evaluation measurable~\citep{lau2018vqarad,liu2021slake,chen2024gmaimmbench}. Volumetric and temporal benchmarks extend evaluation toward 3D medical imaging, CT volume report modeling, or longitudinal CT reasoning~\citep{bai2024m3d,hamamci2024ctrate,tang20253drad}, and brain imaging
benchmarks address multi-stage clinical workflows~\citep{peng2025omnibrainbench}, anomaly localization~\citep{bercea2025novabenchmarkanomalylocalization}, 2D to 3D bridging~\citep{wang2024brainmd}, or cross-sectional brain MRI report generation~\citep{lei2024radgenomebrain}.However, real brain MRI interpretation requires joint reasoning over longitudinal change, multi-sequence findings, volumetric anatomy, and report-grounded clinical evidence (as shown in Figure~\ref{fig:bench_draw}). This leaves open a distinct evaluation question:

\begin{tcolorbox}[colback=blue!3,
  colframe=blue!50!black,
  boxrule=0.5pt,
  arc=1mm,
  left=1.2mm,right=1.2mm,
  top=0.9mm,bottom=0.9mm]
\textbf{Evaluation Question:} Can a VLM trace evidence through longitudinal brain MRI follow-up studies while leveraging multi-sequence MRI cues and volumetric spatial context to perform real clinical reasoning? 
\end{tcolorbox}

Answering this question requires a \emph{report-grounded} benchmark that evaluates four linked demands, including \emph{longitudinal} comparison across
serial studies, \emph{multi-sequence} MRI integration, \emph{volumetric} localization and burden, and \emph{clinical reasoning} beyond surface
description.

To close this gap, we introduce \textbf{BrainTRACE}, a report-grounded brain MRI VQA benchmark containing 7{,}273 scored instances derived from 1{,}778 longitudinal patients, 7{,}299 MRI studies, and approximately 29k co-registered 3D MRI sequence volumes. BrainTRACE is designed around follow-up interpretation rather than generic image question answering. Each instance separates model-facing evidence from hidden clinical references so that models receive only the permitted visual evidence, question, and answer format, while radiology reports and
segmentation-derived measurements define reference answers, required facts, and clinically material contradictions. This design allows the benchmark to evaluate whether a model’s answer is supported by available studies, not merely whether it is plausible.

BrainTRACE then localizes failures through two complementary annotations. The \emph{clinical reasoning depth} axis follows the workflow of longitudinal brain MRI interpretation, from acquisition recognition and single study characterization to interval comparison, trajectory tracking, and case-level synthesis. The \emph{evidence demand} axis records what support is required at each step: longitudinal comparison, multi-sequence integration, volumetric spatial evidence, or clinical reasoning. Together, this separation makes the benchmark more diagnostic so that errors can be interpreted as failures of the requested operation, failures to use the required evidence, or hallucinated facts unsupported by the record. Our contributions can be summarized as follows:

\begin{itemize}
    \item We formalize brain MRI VLM evaluation as longitudinal evidence tracing, where clinical validity depends on comparing serial studies and grounding conclusions in the relevant imaging and clinical references.
    \item We instantiate this definition as BrainTRACE, a 7{,}273 instance benchmark with hidden report and measurement derived references, diagnostic two-axis annotation, three evaluation tracks, exact closed-form scoring, and factual audit scoring for open-ended and case-level answers.
    \item We evaluate 20 model configurations covering general, medical, frontier, and native volume VLMs, and identify a longitudinal evidence composition gap: the best broad compatible closed-form accuracy is 41.9\%, but the strongest open-ended clinical pass is
    19.8\%, and the best case success rate is 1.4\%.
\end{itemize}

%% file: sections/02_related_work.tex
\section{Related Work}
\label{sec:related-work}
\subsection{Medical VQA Benchmarks}

\begin{table}[t]
\centering
\scriptsize
\setlength{\tabcolsep}{1.5pt}
\renewcommand{\arraystretch}{1.08}
\caption{Positioning representative medical VLM/VQA benchmarks relative to BrainTRACE along four evidence attributes. \emph{Longitudinal}: longitudinal or temporal visual evidence. \emph{Multi-seq.}: item-level multi-sequence evidence. \emph{Volumetric}: native-volume or explicit volumetric evidence. \emph{Clinical}: reasoning depth beyond visual recognition. \protect\bvfull{}\,=\,full, \protect\bvpartial{}\,=\,partial, \protect\bvnone{}\,=\,absent.}
\label{tab:gap-teaser}
\begin{tabular*}{\textwidth}{@{\extracolsep{\fill}}llcccc@{}}
\toprule
Benchmark & Domain & \textbf{Longitudinal} & \textbf{Multi-seq.} &
\textbf{Volumetric} & \textbf{Clinical} \\
\midrule
VQA-RAD~\citep{lau2018vqarad}
& Mixed 2D radiology & \bvnone & \bvnone & \bvnone & \bvpartial \\
SLAKE~\citep{liu2021slake}
& Mixed 2D medical imaging & \bvnone & \bvnone & \bvnone & \bvpartial \\
Medical-Diff-VQA~\citep{hu2023medicaldiffvqa}
& Paired chest X-rays & \bvfull & \bvnone & \bvnone & \bvfull \\
GMAI-MMBench~\citep{chen2024gmaimmbench}
& Broad medical imaging & \bvnone & \bvpartial & \bvpartial & \bvfull \\
M3D-Bench~\citep{bai2024m3d}
& 3D medical imaging & \bvnone & \bvnone & \bvfull & \bvpartial \\
3D-RAD~\citep{tang20253drad}
& Longitudinal 3D CT & \bvfull & \bvnone & \bvfull & \bvfull \\
BrainMD~\citep{wang2024brainmd}
& Brain MRI & \bvnone & \bvpartial & \bvpartial & \bvpartial \\
OmniBrainBench~\citep{peng2025omnibrainbench}
& Broad brain imaging & \bvnone & \bvpartial & \bvpartial & \bvfull \\
\midrule
\textbf{BrainTRACE}
& \textbf{Longitudinal brain MRI}
& \bvfull & \bvfull & \bvfull & \bvfull \\
\bottomrule
\end{tabular*}
\vspace{-3mm}
\end{table}


Medical VQA benchmarks have expanded from selected 2D radiology images to
broader, volumetric, temporal, and reasoning-oriented evaluation. VQA-RAD
~\citep{lau2018vqarad} and SLAKE~\citep{liu2021slake} established
radiology VQA on selected 2D medical images, while GMAI-MMBench
~\citep{chen2024gmaimmbench} broadened evaluation across departments,
modalities, and task types. Medical-Diff-VQA~\citep{hu2023medicaldiffvqa}
tests paired chest-X-ray difference reasoning, and CXReasonBench
~\citep{lee2026cxreasonbench} evaluates structured diagnostic reasoning
in chest X-rays. M3D-Bench~\citep{bai2024m3d} evaluates 3D medical image
understanding, CT-RATE~\citep{hamamci2024ctrate} pairs CT volumes with
radiology text, SGMRI-VQA~\citep{moukheiber2026beyond} introduces
frame indexed spatial grounding for volumetric MRI, and 3D-RAD
~\citep{tang20253drad} adds longitudinal temporal diagnosis on CT scans. These benchmarks cover important axes of medical VLM evaluation, but they do not define brain MRI evaluation as report-grounded longitudinal evidence tracing. Brain MRI follow-up is sequence, volume, and time dependent: a clinically valid answer may require comparing serial studies, integrating sequence-specific findings, localizing volumetric burden, and preserving that evidence through clinical synthesis.


\subsection{Brain Imaging Benchmarks and Resources}
Brain imaging benchmarks address brain data more directly, but they still do not make longitudinal brain MRI follow-up the central unit of report-grounded VLM evaluation. OmniBrainBench
~\citep{peng2025omnibrainbench} evaluates multimodal brain imaging
comprehension clinical tasks, largely through 2D visual
evidence. NOVA~\citep{bercea2025novabenchmarkanomalylocalization}
stress tests out-of-distribution anomaly localization across rare brain MRI pathologies. BrainMD~\citep{wang2024brainmd} bridges 2D VLMs and 3D brain MRI through slice selection, while AutoRG-Brain/RadGenome-Brain ~\citep{lei2024radgenomebrain} grounds cross-sectional brain MRI report generation. Complementary resources provide important but different signals: BraTS~\citep{menze2015brats} established a multi-contrast tumor
segmentation, OASIS-3~\citep{lamontagne2019oasis3} provides longitudinal neuroimaging cohorts, and longitudinal MRI methods such as Reuter et al.
~\citep{reuter2012longitudinal} motivates registration-aware serial
comparison.

BrainTRACE differs by making longitudinal brain MRI follow-up the central unit of VLM evaluation. It jointly tests report-grounded comparison across longitudinal studies, multi-sequence MRI integration, volumetric localization, burden assessment, and clinical reasoning beyond surface description.
Table~\ref{tab:gap-teaser} summarizes benchmark-level coverage of these evidence attributes. Broader datasets and non-benchmark resources are included in Appendix~\ref{app:gap-table}.

%% file: sections/03_benchmark.tex
\section{BrainTRACE Design and Construction}
\label{sec:benchmark}

\subsection{From clinical records to evaluation instances}
\label{sec:evaluation-units}

BrainTRACE treats each VQA item as a report-grounded clinical evidence
test rather than a generic image question pair. Each item is derived from
a \textit{longitudinal brain MRI record} containing serial studies,
available MRI sequences, radiology reports, and segmentation-derived
measurements. A key design is to decouple \textit{what the model is allowed to observe} from \textit{what defines clinical correctness}. Concretely, each released instance is organized into three components:
\begin{itemize}
    \item \textbf{Model-facing input:} the permitted visual evidence, the
    question, the required answer format, and allowed non-leaking metadata.
    \item \textbf{Hidden reference:} the report or measurement derived answer target, required facts, accepted variants, and clinical material contradictions used for scoring.
    \item \textbf{Diagnostic labels:} the clinical reasoning depth,
    evidence demand tag, and evaluation track used to localize failures.
\end{itemize}
For closed-form questions, the hidden reference is a normalized target
answer. For open-ended and case-level questions, it further specifies
the facts that must be present and the contradictions that would make an
answer clinically unacceptable. Reports and measurements determine what is clinically supported, while models receive only the permitted visual evidence.



\subsection{Tasks, evidence demands, and evaluation tracks}
\label{sec:tasks}
\label{sec:attributes}

Each BrainTRACE item is annotated along three diagnostic dimensions: what
clinical operation is being tested, what evidence is required, and under which input condition the item is evaluated. This separation is central to the benchmark: a failure may reflect the wrong clinical operation, missing evidence use, or an unsupported clinical synthesis.

\begin{figure}[ht]
\centering
\includegraphics[width=\linewidth]{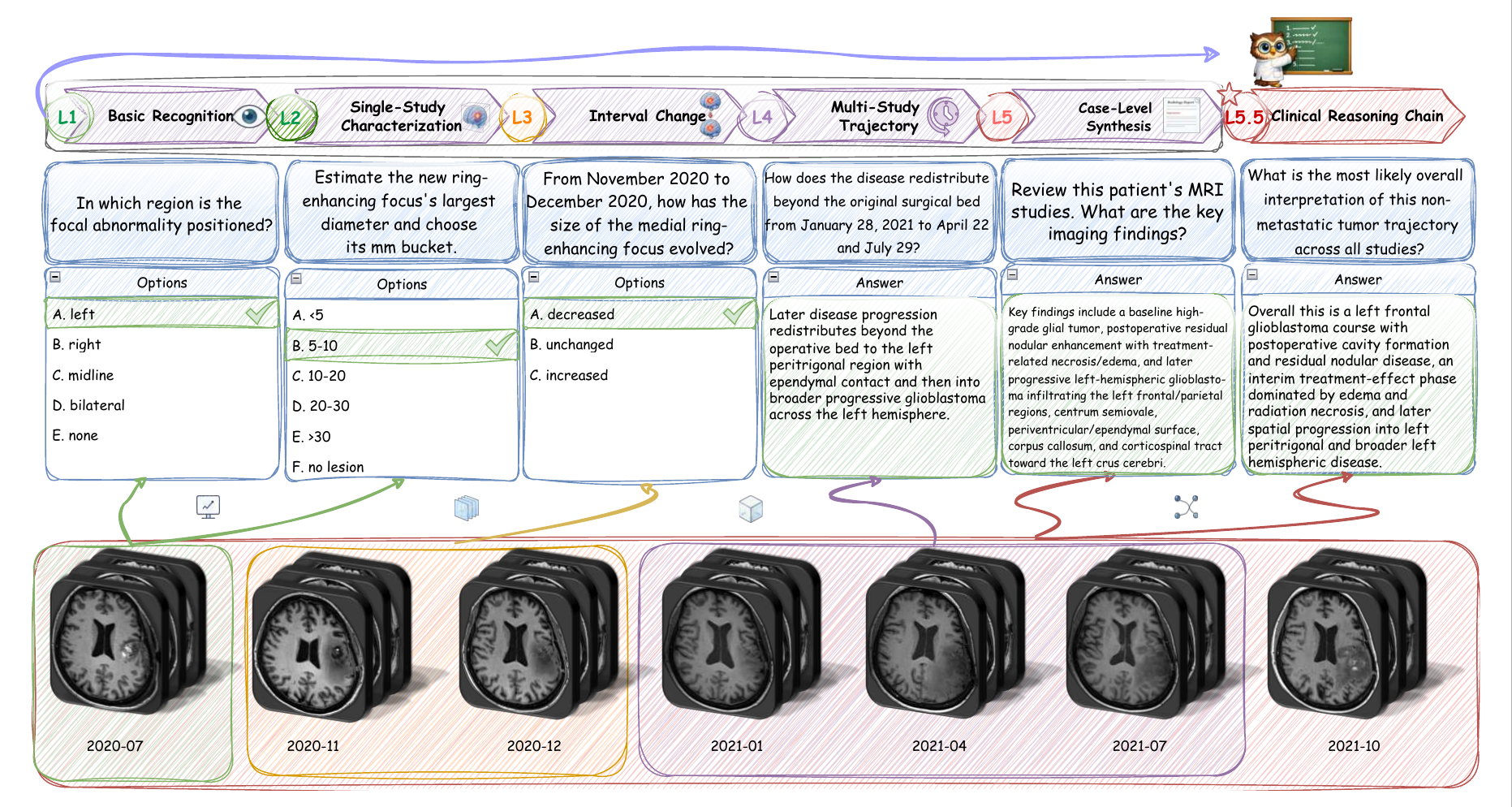}
\caption{A running longitudinal case illustrates how BrainTRACE turns one
patient record into localization, interval comparison, trajectory, and
case synthesis tasks. The case reasoning track further decomposes the
longitudinal interpretation into six report-grounded VQA steps.}
\label{fig:case-reasoning-example}
\end{figure}

\paragraph{Clinical reasoning depth.} We first instantiate the clinical reasoning depth axis as a five-level task taxonomy. L1 covers acquisition and basic detection. L2 covers single study characterization, including burden, location, and appearance. L3 tests interval comparison and disease-specific effects. L4 tests trajectory, timing, and treatment response. L5 covers case-level synthesis. This taxonomy makes failures clinically
localizable: an answer can fail because the model missed the abnormality,
compared the wrong timepoints, misread burden evolution, or produced a
synthesis unsupported by the available evidence. Appendix \ref{app:clinical-taxonomy} defines the clinical reasoning depth and example
question families. Table~\ref{tab:tmpl-props} in Appendix gives the
per-track denominators and task family inventory.

\paragraph{Evidence demand.}
The evidence demands tag records what support is required for a valid
answer. Longitudinal questions require comparison across time points from
the same patient. Multi-sequence questions require complementary MRI
sequences within a study, such as FLAIR abnormality, enhancement,
susceptibility, edema, or mass effect. Volumetric questions require
spatial evidence beyond a single slice, either through multi-slice
rendering or supporting native-volume input. Clinical reasoning questions
require interpretation beyond surface recognition, such as treatment
response, disease status, disease burden, or differential diagnosis.
These tags are orthogonal to the operation levels: the same operation may
require different forms of evidence, and the same evidence demand may
appear across multiple levels. Table~\ref{tab:tmpl-props} in Appendix lists
the demand tags for each task family.


\paragraph{Evaluation tracks.}
BrainTRACE instantiates these annotations through three evaluation tracks. The broad compatible track contains 5{,}653 instances and renders evidence into standard image inputs, allowing both open-weight and closed-source VLMs to be evaluated under the same interface. The 3D-evidence track contains 1{,}200 items and evaluates volumetric clinical targets using rendered multi-slice evidence or supported native-volume inputs. The case reasoning track contains 70 longitudinal cases and 420 decomposed VQA steps. It makes the longitudinal evidence chain explicit (as shown in Figure~\ref{fig:case-reasoning-example}) by decomposing each case into six report-grounded VQA steps: anchor
localization, early interval change, interval characterization, burden
trajectory, latest status, and final synthesis. This decomposition turns
an opaque case-level failure into a localized failure along the evidence
path.

\subsection{Report-grounded benchmark construction}
\label{sec:cohort}

\begin{figure}[ht]
\centering
\includegraphics[width=\linewidth]{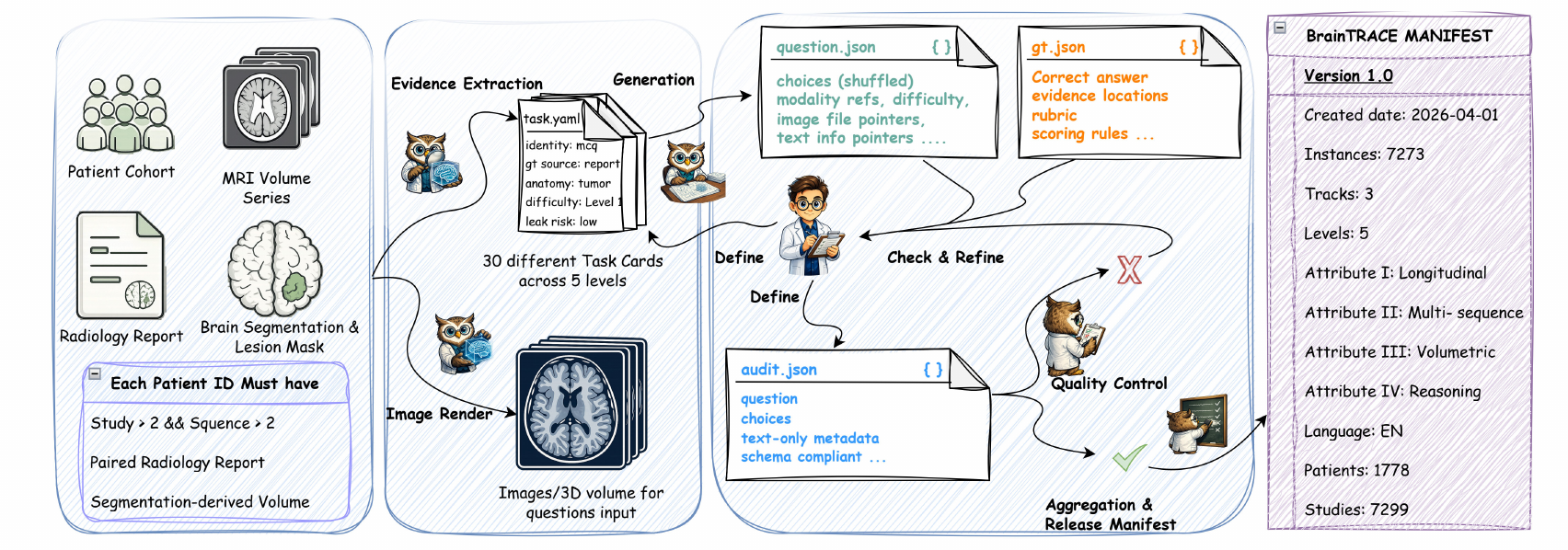}
\caption{BrainTRACE constructs report-grounded VQA instances through a
human-in-the-loop pipeline that separates hidden clinical references from
model-facing visual evidence. Reports and segmentation-derived measurements
define answer targets, required facts, contradictions, and scoring rubrics.
Models receive only the permitted visual inputs.}
\label{fig:benchmark-pipeline}
\end{figure}

BrainTRACE is derived from the publicly released MR-RATE
corpus~\citep{mrrate2026}. We retain patients with sufficient
longitudinal context for evidence chain questions: each retained patient
has at least three MRI studies with paired imaging, radiology reports,
and segmentation-derived anatomical measurements. Each retained study is
centered on a T1-weighted acquisition and may include co-registered
T2-weighted, FLAIR, SWI, or other clinically relevant sequences. The
benchmark targets longitudinal brain MRI follow-up studies from
MR-RATE, rather than non-longitudinal referrals or the full
neuroradiology spectrum. Appendix~\ref{app:cohort-composition} reports
the upstream corpus scale and retained cohort composition.

The clinical references come from authentic MR-RATE radiology reports
rather than synthetic report text. In the MR-RATE release, reports are
anonymized and structured into clinical information, technique, findings,
and impression fields through an iterative LLM-assisted construction
procedure. BrainTRACE uses these structured reports as source clinical
documents, then links report-supported findings to available imaging
studies and segmentation-derived measurements.

Figure~\ref{fig:benchmark-pipeline} summarizes the human-in-the-loop
construction pipeline. The process first identifies report-supported
evidence statements, including findings, anatomical locations, interval relations across studies, disease burden statements, disease status
language, and quantitative measurements. Template eligibility rules then
decide which statements can support a task: for example, an interval change
question requires a finding described at comparable timepoints, whereas a
measurement question requires a segmentation-derived value aligned to the
relevant study. Eligible evidence is converted into a task card containing
the question, answer format, hidden ground truth target or rubric,
diagnostic labels, and the rendered visual input under the specified
evidence condition. Human reviewers inspect the task cards and
intermediate outputs throughout this process, checking report support,
timepoint alignment, visual evidence availability, answer format validity,
redaction, and scoring consistency. Appendix
Section~\ref{app:construction-artifacts} describes the model-facing and
non-model-facing records used to reconstruct and audit these steps.

\subsection{Quality Control}
\label{sec:leakage-reproducibility}
\label{sec:groundtruth-qc}

\paragraph{Ground truth control.} BrainTRACE stores a non-model-facing hidden reference for each item, including the target answer, required supporting facts, clinically material contradictions, and scoring criteria. Closed-form items use a lightly normalized target answer. Open-ended items use a rubric of required facts, optional facts, and record contradicting statements. For case reasoning items, each decomposed VQA step has its own reference, so the evaluation can identify where the evidence path breaks. The case reasoning track receives item level review by an American Board of
Radiology (ABR) certified radiation oncologist, including the question,
ground truth, and audit rubric for each decomposed step.

\paragraph{Leakage control.} BrainTRACE controls leakage by separating inputs from hidden clinical records. During evaluation, models receive only the rendered visual evidence, the question, the answer format, and allowed non-leaking metadata. They do not receive radiology reports, parsed report fields, segmentation masks, measurement tables, or hidden reference fields.
Metadata is redacted when it could reveal the answer. Each instance
passes quality control checks for redaction, report support, visual
evidence availability, and scoring consistency. We also run text-only
leakage probes to test whether the question wording alone is sufficient
to infer the answer without visual evidence.


%% file: sections/04_experiments.tex
\section{Experiments}
\label{sec:experiments}

\begin{table}[t]
\centering
\footnotesize
\setlength{\tabcolsep}{0.9pt}
\begin{threeparttable}
\caption{Closed-form accuracy (\%) is fragmented across clinical subclasses. Subclass columns pool broad-compatible and 3D-evidence items. Overall columns separate tracks.}
\label{tab:closed-diagnostic}
\begin{tabular*}{\textwidth}{@{\extracolsep{\fill}}lccccccccccccc@{}}
\toprule
 & \multicolumn{2}{c}{\textbf{L1 recognition}} &
\multicolumn{3}{c}{\textbf{L2 single-study}} &
\multicolumn{3}{c}{\textbf{L3 interval}} &
\multicolumn{3}{c}{\textbf{L4 trajectory}} &
\multicolumn{2}{c}{\textbf{Overall}} \\
\cmidrule(lr){2-3}\cmidrule(lr){4-6}\cmidrule(lr){7-9}\cmidrule(lr){10-12}\cmidrule(lr){13-14}
\textbf{Model} & \textbf{ACQ} & \textbf{DET} & \textbf{BUR} & \textbf{LOC} & \textbf{APP} &
\textbf{QNT} & \textbf{EVT} & \textbf{EFF} & \textbf{TRJ} & \textbf{TMP} & \textbf{RSP} & \textbf{Broad} & \textbf{3D} \\
\midrule
\bvgroup{14}{BVGeneralBand}{Open-weight general VLMs}
\qwenicon Qwen2.5-VL-7B~\citep{bai2025qwen25vl} & 47.5 & 38.0 & 7.1 & 44.9 & 27.8 & 26.8 & \bvranksecond{53.7} & 26.2 & 25.6 & 20.3 & 38.5 & 33.2 & 33.2 \\
\qwenicon Qwen3-VL-4B~\citep{bai2025qwen3vl} & 47.2 & 41.3 & 8.7 & 44.6 & 31.4 & \bvranksecond{36.9} & 52.6 & 28.0 & 13.9 & 16.8 & 27.3 & 32.7 & 35.4 \\
\qwenicon Qwen3-VL-8B~\citep{bai2025qwen3vl} & 49.8 & 42.0 & 12.0 & 37.0 & 32.4 & 33.7 & 51.1 & 30.9 & 25.4 & 17.2 & 25.9 & 33.7 & 33.4 \\
\qwenicon Qwen3-VL-30B~\citep{bai2025qwen3vl} & 64.5 & \bvranksecond{44.4} & \bvranksecond{23.2} & 33.5 & 30.4 & \bvrankbest{37.2} & 52.6 & 28.6 & 25.4 & 15.8 & 10.5 & 35.1 & 34.8 \\
\qwenicon Qwen3.5-27B~\citep{qwen2026qwen35} & 81.5 & 38.9 & 13.9 & 28.0 & 30.2 & 23.8 & 35.8 & 43.4 & 24.5 & 14.0 & 18.9 & 37.3 & 12.6 \\
\internvlicon InternVL3-8B~\citep{zhu2025internvl3} & 63.0 & 40.7 & 6.0 & \bvrankbest{48.2} & 38.6 & 34.8 & 49.1 & 29.7 & 24.9 & 15.2 & 25.2 & 35.8 & 35.3 \\
\internvlicon InternVL3-14B~\citep{zhu2025internvl3} & 56.8 & 40.2 & 7.7 & 42.9 & 37.4 & 31.0 & 53.1 & 27.7 & 29.3 & 18.6 & 16.1 & 35.1 & 34.0 \\
\internvlicon InternVL3-78B~\citep{zhu2025internvl3} & 60.5 & 40.2 & 10.7 & 43.2 & 35.0 & 25.9 & 53.6 & 30.3 & 24.5 & 19.4 & 35.7 & 35.5 & 33.3 \\
\deepseekicon Janus-Pro-7B~\citep{chen2025januspro} & 1.0 & 34.2 & 1.1 & 3.2 & 0.6 & 3.8 & 44.8 & 0.3 & 7.5 & 15.1 & 4.2 & 10.9 & 17.6 \\
\midrule
\bvgroup{14}{BVMedicalBand}{Open-weight medical VLMs}
\huatuoicon HuatuoGPT-Vision-7B~\citep{chen2024huatuogptvision} & 76.2 & 38.7 & 7.1 & 44.3 & 31.6 & 25.6 & 47.9 & 25.9 & 26.8 & 18.6 & 34.3 & 36.3 & 29.1 \\
\lingshuicon Lingshu-7B~\citep{xu2025lingshu} & 78.8 & 37.8 & 7.9 & \bvranksecond{46.3} & 27.6 & 30.4 & \bvrankbest{54.2} & 29.7 & 26.2 & 19.5 & 30.1 & 36.5 & 35.4 \\
\gemmaicon MedGemma-4B-it~\citep{sellergren2025medgemma} & 42.2 & 35.5 & 2.7 & 20.2 & 28.2 & 7.5 & 47.8 & 25.4 & 15.2 & 14.5 & 0.7 & 24.9 & 18.9 \\
\llavaicon LLaVA-Med-v1.5~\citep{li2023llavamed} & 0.0 & 2.9 & 0.0 & 0.2 & 0.4 & 0.0 & 0.7 & 0.0 & 0.0 & 0.0 & 0.0 & 0.6 & 0.0 \\
\midrule
\bvgroup{14}{BVClosedBand}{Closed-source frontier VLMs}
\openaiicon GPT-5.4~\citep{openai2026gpt54} & \bvranksecond{87.5} & 42.2 & 15.3 & 37.9 & \bvrankbest{40.6} & 35.1 & 50.2 & \bvranksecond{49.6} & \bvranksecond{32.0} & \bvranksecond{23.1} & \bvranksecond{44.1} & \bvoverallsecond{40.4} & \bvoverallbest{42.2} \\
\openaiicon GPT-5~\citep{openai2026gpt54} & \bvrankbest{88.8} & 40.5 & 10.7 & 40.5 & \bvranksecond{39.0} & 31.5 & 51.9 & \bvrankbest{58.3} & \bvrankbest{33.9} & 21.7 & \bvrankbest{49.7} & \bvoverallbest{41.9} & \bvoverallsecond{38.3} \\
\openaiicon GPT-5-mini~\citep{openai2026gpt54} & 84.2 & 42.4 & \bvrankbest{25.2} & 30.7 & 34.4 & 28.0 & 50.4 & 49.3 & 27.7 & 19.0 & 42.7 & 38.2 & 37.5 \\
\anthropicicon Claude Opus 4.6~\citep{anthropic2026claudeopus46} & 79.2 & 14.2 & 8.5 & 30.9 & 22.8 & 12.0 & 9.2 & 33.5 & 20.6 & 20.4 & 18.9 & 29.8 & 1.2 \\
\geminiicon Gemini 2.5 Pro~\citep{googledeepmind2025gemini25pro} & 85.2 & \bvrankbest{49.3} & 19.5 & 21.7 & 29.8 & 29.6 & 48.3 & 34.4 & 26.6 & \bvrankbest{24.0} & 14.7 & 36.5 & 32.0 \\
\midrule
\bvgroup{14}{BVNativeBand}{Native-volume medical baselines}
\baaiicon M3D-LaMed~\citep{bai2024m3d} & -- & 36.7 & 18.5 & 40.4 & 30.4 & 24.0 & 51.7 & 14.3 & 18.4 & 17.1 & 26.1 & -- & 29.9 \\
\radfmicon RadFM~\citep{wu2025radfm} & 33.8 & 28.2 & 8.2 & 9.9 & 16.4 & 16.0 & 28.3 & 5.5 & 12.3 & 16.5 & 2.1 & 18.1 & 15.0 \\
\midrule
$n$ & 400 & 550 & 366 & 657 & 500 & 575 & 601 & 344 & 482 & 558 & 143 & 4{,}053 & 1{,}123 \\
\bottomrule
\end{tabular*}
\begin{tablenotes}[flushleft]
\scriptsize
\item ACQ=acquisition, DET=detection/localization, BUR=burden, LOC=location, APP=appearance, QNT=quantitative change, EVT=lesion event, EFF=structural effect, TRJ=trajectory, TMP=timing, RSP=response. Best values are bold. Second-best values are underlined. Only the Overall columns use background shading.
\end{tablenotes}
\end{threeparttable}
\vspace{-3mm}
\end{table}

\subsection{Experimental Setup}
\label{sec:experiment-setup}

We evaluate 20 model configurations spanning open-weight general VLMs,
open-weight medical VLMs, closed-source frontier VLMs, and available
native-volume medical baselines (Table~\ref{tab:closed-diagnostic}). All
models are evaluated zero-shot using only the permitted model-facing
visual evidence, question, answer format, and allowed non-leaking
metadata. Reports, segmentation masks, parsed findings, measurements, and
hidden reference fields are never shown. Each model receives one attempt
per item. Malformed outputs, refusals, and abstentions remain in the
denominator. For 3D evidence items, general VLMs use released rendered
inputs, while native-volume medical baselines use their supported
volume input interfaces. Thus, 3D evidence results compare supported
interfaces on shared clinical targets rather than isolating native-volume
representation as a controlled variable.

\subsection{Evaluation Metrics}
\label{sec:evaluation-metrics}

\paragraph{Closed-form accuracy.}
Closed-form questions are scored by exact match after light answer
normalization and uniform removal of visible inline reasoning blocks. We
report subclass accuracy by clinical reasoning depth and overall accuracy
separately for the broad compatible and 3D evidence tracks.
\paragraph{Open-ended and case-level clinical pass.}
Open-ended answers and decomposed case-level steps are scored by a fixed
factual clinical audit. The audit judge receives the question, the
model's final answer, and the item-level rubric, including the reference
answer, required clinical facts, critical slots, accepted paraphrases,
and auto fail conditions. It does not receive the source report,
segmentation tables, evidence quotes, or hidden reference record. 

The audit measures factual clinical acceptability rather than prose
quality. Let $F_i$, $S_i$, $E_i$, and $U_i$ denote required fact coverage,
critical slot coverage, absence of critical error, and absence of a
clinically material unsupported claim. The item-level pass indicator is $P_i = F_i S_i E_i U_i$.
For the case reasoning track, Step Pass averages $P_i$ over decomposed
steps, while Case Success requires all six steps in the same longitudinal
case to pass: $\mathrm{CaseSuccess}_c = \prod_{s=1}^{6} P_{c,s}$.
Appendix~\ref{app:open-scoring-protocol} details the scoring fields,
derived metrics, edge case policy, and released audit records.
Appendix~\ref{appx:iaa} and Appendix~\ref{appx:judge-family-sensitivity}
report agreement and judge family sensitivity analyses.

\subsection{Experimental Results}
\label{sec:experimental-results}


\paragraph{Closed-form accuracy is fragmented across clinical skills.}
Table~\ref{tab:closed-diagnostic} reports closed-form accuracy across
clinical subclasses. The strongest frontier models cluster near 40\%
broad compatible accuracy: GPT-5 obtains the best broad score at 41.9\%,
and GPT-5.4 obtains the best 3D evidence score at 42.2\%. However, the
aggregate score hides large variation across acquisition recognition,
burden, localization, interval change, trajectory, timing, and response
assessment. Closed-form accuracy indicates partial success on
isolated evidence tests, not reliable use of disease burden, temporal
change, or treatment response evidence.


\input{sections/B_appendix_open_case_table}

\paragraph{Open-ended clinical acceptability remains much lower than
recognition accuracy.}
Table~\ref{tab:open-v06} shows that factual clinical pass is substantially
harder than closed-form recognition. GPT-5.4 is strongest on this axis,
but passes only 19.8\% of broad open-ended questions and 15.6\% of
3D evidence open-ended questions. Failures are clinically material:
models often mention a relevant abnormality but omit disease status,
reverse interval direction, assign evidence to the wrong time point, or add unsupported anatomy. As a clinical answerability sanity check, one board-certified professional confirms that the selected cases are clinically answerable under expert reading, but it is not a directly comparable model baseline. Appendix~\ref{app:open-leaderboards} reports the full open-ended audit leaderboard and secondary text metrics, while Appendix~\ref{app:case-examples} gives paired success/failure examples that make the factual audit failure modes inspectable.

\paragraph{Case-level reasoning fails by evidence attrition.}

\begin{figure}[t]
\centering
\includegraphics[width=\linewidth]{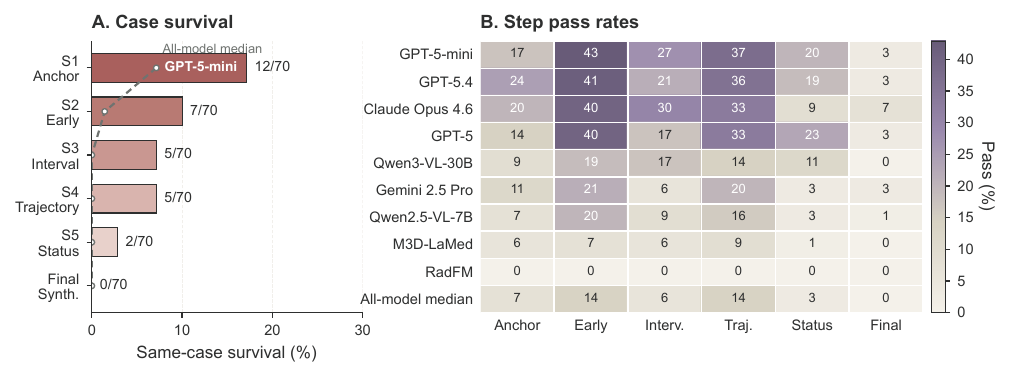}
\caption{Case-level reasoning fails by attrition. Local step successes
rarely survive through all six decomposed VQA steps in the same
longitudinal case.}
\label{fig:case-attrition}
\end{figure}

The case-reasoning results in Table~\ref{tab:open-v06} and
Figure~\ref{fig:case-attrition} exposes a stronger compositional failure.
The best aggregate Step Pass is 24.5\%, but this overstates the case-level
ability because different steps often pass in different cases. Only two
model rows complete any case, both at 1.4\% Case Success. Models may
identify an anchor abnormality or describe one interval, but these local
successes rarely survive through the latest status interpretation and final
synthesis. Local perception and interval reasoning are necessary but not sufficient for report-grounded longitudinal
interpretation. Appendix~\ref{app:case-reasoning-audit} provides the released per-step and per-case audit records, and Appendix~\ref{app:case-step-attrition} further decomposes the attrition pattern in Figure~\ref{fig:case-attrition}.


\paragraph{Evidence-channel ablations confirm dependence on retained
evidence.}
\label{sec:evidence-ablation}

Table~\ref{tab:evidence-channel-ablation} reports matched diagnostic
ablations for removing longitudinal evidence by
using only the latest timepoint drops accuracy by 37.5 points for
Qwen3.5-27B and 23.3 points for GPT-5.4. Restricting inputs to a single
T2 sequence also reduces performance, and removing images entirely causes
large drops for both models. 
The drops show that performance is not explained only by answer priors or question wording. It depends on retained longitudinal, multi-sequence, and visual evidence. Appendix~\ref{app:evidence-ablation} details the matched
evidence channel ablations, and Appendix~\ref{app:empirical-diagnostics}
reports a complementary 3D interface, rank sensitivity, and chance calibrated
difficulty diagnostics.

\begin{table}[th]
\centering
\small
\setlength{\tabcolsep}{4.5pt}
\renewcommand{\arraystretch}{1.08}
\caption{Evidence-channel ablations reduce accuracy on matched
diagnostic items. All ablation rows use $n=120$ with a constructed
50.0\% full-evidence baseline, so drops are comparable across
conditions, but conditional on the balanced sample.}
\label{tab:evidence-channel-ablation}
\begin{tabular}{lccc|rrrr}
\toprule
Condition & \multicolumn{3}{c|}{Evidence retained} &
\multicolumn{2}{c}{Accuracy (\%)} & \multicolumn{2}{c}{Drop (pp)} \\
\cmidrule(lr){2-4}\cmidrule(lr){5-6}\cmidrule(lr){7-8}
 & Long. & Multi-seq. & Image & Qwen & GPT & Qwen & GPT \\
\midrule
Full evidence & \tabcheck & \tabcheck & \tabcheck
& \textbf{50.0} & \textbf{50.0} & -- & -- \\
Latest timepoint only & -- & \tabcheck & \tabcheck
& 12.5 & 26.7 & -37.5 & -23.3 \\
T2 sequence only & \tabcheck & -- & \tabcheck
& 29.2 & 38.3 & -20.8 & -11.7 \\
No image & -- & -- & --
& 15.0 & 9.2 & -35.0 & -40.8 \\
\bottomrule
\end{tabular}
\end{table}

%% file: sections/B_appendix_open_case_table.tex
\begin{table}[t]
\centering
\footnotesize
\setlength{\tabcolsep}{0.8pt}
\begin{threeparttable}
\caption{Open-ended acceptability and case-reasoning remain low
even for frontier models; entries are Pass \%.}
\label{tab:open-v06}
\begin{tabular*}{\textwidth}{@{\extracolsep{\fill}}lccccccccccc@{}}
\toprule
& \multicolumn{4}{c}{\textbf{Open subclasses}} &
\multicolumn{3}{c}{\textbf{Case synthesis}} &
\multicolumn{2}{c}{\textbf{Overall}} &
\multicolumn{2}{c}{\textbf{Case reasoning}} \\
\cmidrule(lr){2-5}\cmidrule(lr){6-8}\cmidrule(lr){9-10}\cmidrule(lr){11-12}
\textbf{Model} & \textbf{L1-ABN} & \textbf{L2-APP} & \textbf{L3-EFF} & \textbf{L4-TRJ} & \textbf{IMP} & \textbf{CMP} & \textbf{DIF} & \textbf{Broad} & \textbf{3D} & \textbf{Step} & \textbf{Case} \\
\midrule
\faUserMd~ABR diplomate oncologist & 100 & 90 & 80 & 100 & 100 & 100 & 100 & 92.9 & 94.3 &91.7 & 80 \\
\bvgroup{12}{BVGeneralBand}{Open-weight general VLMs}
\qwenicon Qwen2.5-VL-7B~\citep{bai2025qwen25vl} & 12.0 & 8.5 & 7.6 & 1.4 & 2.3 & 2.8 & 2.3 & 4.9 & 3.9 & 9.3 & \bvoverallbest{1.4} \\
\qwenicon Qwen3-VL-4B~\citep{bai2025qwen3vl} & 15.0 & 15.5 & 17.5 & 5.2 & 9.4 & 10.8 & 9.4 & 11.8 & 5.2 & 8.3 & 0.0 \\
\qwenicon Qwen3-VL-8B~\citep{bai2025qwen3vl} & 11.5 & 18.0 & \bvranksecond{21.3} & 6.6 & 11.7 & 24.9 & 2.8 & 13.8 & 9.1 & 3.8 & 0.0 \\
\qwenicon Qwen3-VL-30B~\citep{bai2025qwen3vl} & 14.0 & 9.5 & 10.9 & 6.2 & \bvranksecond{12.7} & \bvrankbest{32.9} & \bvrankbest{15.0} & \bvoverallsecond{14.4} & 11.7 & \bvoverallsecond{11.7} & 0.0 \\
\qwenicon Qwen3.5-27B~\citep{qwen2026qwen35} & 17.5 & 7.5 & 10.0 & 0.9 & 7.5 & 9.4 & 3.3 & 8.1 & 3.9 & 2.9 & 0.0 \\
\internvlicon InternVL3-8B~\citep{zhu2025internvl3} & 17.0 & 10.5 & 3.5 & 3.0 & 3.8 & 6.0 & 5.0 & 6.6 & -- & 8.3 & 0.0 \\
\internvlicon InternVL3-14B~\citep{zhu2025internvl3} & 9.0 & 14.0 & 11.0 & 1.0 & 8.3 & 9.5 & 7.5 & 8.6 & -- & 11.2 & 0.0 \\
\internvlicon InternVL3-78B~\citep{zhu2025internvl3} & 11.0 & 18.5 & 7.5 & 2.5 & 4.8 & 6.0 & 5.5 & 7.6 & -- & 0.0 & 0.0 \\
\deepseekicon Janus-Pro-7B~\citep{chen2025januspro} & 3.5 & 7.5 & 0.9 & 0.0 & 0.0 & 0.0 & 2.3 & 1.8 & 0.0 & 0.5 & 0.0 \\
\midrule
\bvgroup{12}{BVMedicalBand}{Open-weight medical VLMs}
\huatuoicon HuatuoGPT-Vision-7B~\citep{chen2024huatuogptvision} & \bvranksecond{23.5} & 15.5 & 4.7 & \bvranksecond{7.6} & 1.9 & 7.5 & \bvranksecond{14.1} & 9.9 & 0.0 & 0.2 & 0.0 \\
\lingshuicon Lingshu-7B~\citep{xu2025lingshu} & 20.5 & 10.5 & 8.5 & 1.9 & 3.1 & 1.4 & 5.6 & 6.8 & 3.9 & 6.9 & 0.0 \\
\gemmaicon MedGemma-4B-it~\citep{sellergren2025medgemma} & 15.5 & 17.0 & 4.3 & 0.0 & 0.0 & 0.5 & 6.6 & 5.4 & 3.9 & 7.6 & 0.0 \\
\llavaicon LLaVA-Med-v1.5~\citep{li2023llavamed} & 4.0 & \bvranksecond{19.0} & 0.9 & 0.5 & 0.5 & 2.8 & 1.9 & 3.8 & 1.3 & 0.2 & 0.0 \\
\midrule
\bvgroup{12}{BVClosedBand}{Closed-source frontier VLMs}
\openaiicon GPT-5.4~\citep{openai2026gpt54} & \bvrankbest{24.5} & \bvrankbest{30.0} & \bvrankbest{26.1} & \bvrankbest{17.1} & \bvrankbest{14.6} & 17.8 & 13.6 & \bvoverallbest{19.8} & \bvoverallbest{15.6} & \bvoverallsecond{24.0} & 0.0 \\
\openaiicon GPT-5~\citep{openai2026gpt54} & 15.0 & 10.0 & 19.5 & 0.5 & 12.1 & 12.6 & 2.5 & 10.5 & 11.7 & 21.7 & \bvoverallbest{1.4} \\
\openaiicon GPT-5-mini~\citep{openai2026gpt54} & 16.0 & 6.0 & 12.5 & 3.5 & 11.6 & 22.1 & 1.5 & 10.6 & 10.4 & \bvoverallbest{24.5} & 0.0 \\
\anthropicicon Claude Opus 4.6~\citep{anthropic2026claudeopus46} & 22.0 & 6.0 & 12.5 & 1.0 & 9.8 & 15.1 & 1.0 & 9.6 & 6.5 & 23.1 & 0.0 \\
\geminiicon Gemini 2.5 Pro~\citep{googledeepmind2025gemini25pro} & 20.5 & 15.0 & 16.6 & 2.4 & 10.1 & \bvranksecond{28.6} & 10.3 & 14.2 & \bvoverallsecond{13.0} & 10.7 & 0.0 \\
\midrule
\bvgroup{12}{BVNativeBand}{Native-volume medical baselines}
\baaiicon M3D-LaMed~\citep{bai2024m3d} & -- & -- & 0.0 & 0.0 & 0.0 & 0.0 & 0.0 & -- & 0.0 & 4.8 & 0.0 \\
\radfmicon RadFM~\citep{wu2025radfm} & 6.0 & 13.5 & 4.3 & 1.4 & 1.4 & 4.7 & 1.4 & 4.3 & 2.6 & 0.0 & 0.0 \\
\midrule
$n$ & 200 & 200 & 211 & 211 & 426 & 213 & 213 & 1{,}597 & 77 & 420 & 70 \\
\bottomrule
\end{tabular*}
\begin{tablenotes}[flushleft]
\scriptsize
\item 
APP=appearance description, EFF=interval effect, TRJ=trajectory summary,
IMP=impression, CMP=comparative summary, DIF=differential diagnosis.
Step denotes Step Pass over decomposed VQA steps; Case denotes Case
Success, requiring all six steps in a case to pass. Broad denotes the
broad-compatible track. The physician row is a separate 10-case
image-grounded MCQ sanity check, not a directly comparable model
baseline.
\end{tablenotes}
\end{threeparttable}
\vspace{-3mm}
\end{table}

%% file: sections/06_conclusion.tex
\section{Conclusion}
\label{sec:conclusion}

We introduced \textbf{BrainTRACE}, a report-grounded benchmark for
longitudinal brain MRI VLM evaluation. BrainTRACE asks whether models can
trace evidence across serial studies, MRI sequences, volumetric anatomy,
and report-derived clinical references, rather than answer isolated image
questions. The main finding is not simply that current VLMs score low, but
that their successes do not compose: local recognition rarely survives
through longitudinal, report-grounded clinical synthesis. This gap persists
under current 3D evidence and native-volume interfaces, suggesting that
volumetric access alone is insufficient without stronger multi-sequence,
temporal, and evidence-grounded reasoning. We release the benchmark
specification, evaluation lists, scoring implementation, factual audit protocol, and audit record format to support reproducible evaluation of
future brain MRI VLMs. Limitations and responsible-use considerations are summarized in Appendix~\ref{app:limitations-responsible-use}.

\clearpage

%% file: sections/A_appendix_gap_table.tex

\appendix

\section{Benchmark positioning and scope}
\label{app:gap-table}

This appendix chapter provides the evidence-attribute positioning behind
Section~\ref{sec:related-work}. The purpose is not to rank prior
resources, many of which address important medical VQA, temporal
imaging, 3D imaging, or brain-MRI tasks. Instead, the comparison asks a
narrow benchmark-design question: whether a resource jointly evaluates
report-grounded longitudinal brain-MRI reasoning across longitudinal,
multi-sequence, volumetric, and clinically interpretive evidence.

Table~\ref{tab:gap-full} extends the main-text positioning table
(Table~\ref{tab:gap-teaser}) to additional benchmarks, datasets,
cohorts, and imaging resources considered during benchmark design. The
columns are evidence attributes rather than quality judgments:
\textbf{L} denotes longitudinal or temporal evidence, \textbf{M}
item-level multi-sequence evidence, \textbf{V} volumetric evidence, and
\textbf{C} clinical reasoning beyond surface recognition. Scale is
reported in each source's native unit, such as images, studies, volumes,
questions, or evaluated models; it should not be read as a normalized
dataset-size comparison.

\begin{table}[h]
\centering
\small
\setlength{\tabcolsep}{3pt}
\renewcommand{\arraystretch}{1.08}
\caption{Existing resources cover important parts of medical VQA,
temporal imaging, 3D imaging, and brain-MRI evaluation, but do not
simultaneously instantiate the four evidence attributes targeted by
BrainTRACE. Scale is reported in each source's native unit for
provenance and is not a normalized dataset-size comparison.
\protect\bvfull{}\,=\,full, \protect\bvpartial{}\,=\,partial,
\protect\bvnone{}\,=\,absent.}
\label{tab:gap-full}
\resizebox{\textwidth}{!}{%
\begin{tabular}{@{}llcccccl@{}}
\toprule
Benchmark / resource & Year & Scale & \textbf{L} & \textbf{M} & \textbf{V} & \textbf{C} & Primary task \\
\midrule
\multicolumn{8}{l}{\emph{2D medical VQA / report benchmarks}} \\
VQA-RAD~\citep{lau2018vqarad} & 2018 & 315 imgs / 3.5k QA & \bvnone & \bvnone & \bvnone & \bvnone & 2D VQA \\
PathVQA~\citep{he2020pathvqa} & 2020 & 5k imgs / 33k QA & \bvnone & \bvnone & \bvnone & \bvnone & Pathology VQA \\
SLAKE~\citep{liu2021slake} & 2021 & 642 imgs / 14k QA & \bvnone & \bvnone & \bvnone & \bvnone & Bilingual VQA \\
MIMIC-CXR~\citep{johnson2019mimiccxr} & 2019 & 65k pts / 227k studies & \bvfull & \bvnone & \bvnone & \bvfull & CXR report gen \\
CheXpert~\citep{irvin2019chexpert} & 2019 & 65k pts / 224k imgs & \bvpartial & \bvnone & \bvnone & \bvnone & CXR class. \\
ReXrank / ReXPref-Prior~\citep{zhang2024rexrank} & 2024 & 10k studies & \bvnone & \bvnone & \bvnone & \bvfull & CXR report leaderboard \\
Medical-Diff-VQA~\citep{hu2023medicaldiffvqa} & 2023 & 164k pairs / 700k QA & \bvfull & \bvnone & \bvnone & \bvfull & CXR pair VQA \\
MS-CXR-T~\citep{bannur2023learning} & 2023 & 1.3k pairs + 361 sents & \bvfull & \bvnone & \bvnone & \bvfull & Temporal CXR class. \\
Chest ImaGenome~\citep{wu2021chestimagenome} & 2021 & 242k scene graphs & \bvfull & \bvnone & \bvnone & \bvfull & CXR progression \\
OmniMedVQA~\citep{hu2024omnimedvqa} & 2024 & 118k imgs / 128k QA & \bvnone & \bvnone & \bvpartial & \bvnone & Multi-domain VQA \\
Med-Flamingo eval~\citep{moor2023medflamingo} & 2023 & \textasciitilde{}1k USMLE items & \bvnone & \bvnone & \bvnone & \bvpartial & Few-shot VQA \\
\midrule
\multicolumn{8}{l}{\emph{Text-only MCQA (radiology-adjacent)}} \\
MedQA~\citep{jin2020disease} & 2020 & 12.7k Qs & \bvnone & \bvnone & \bvnone & \bvfull & USMLE MCQA \\
MedMCQA~\citep{pal2022medmcqa} & 2022 & 194k Qs & \bvnone & \bvnone & \bvnone & \bvfull & MCQA (Indian PG) \\
MMLU (clinical)~\citep{hendrycks2021mmlu} & 2021 & \textasciitilde{}1.8k Qs & \bvnone & \bvnone & \bvnone & \bvfull & MCQA \\
\midrule
\multicolumn{8}{l}{\emph{3D and multi-sequence medical imaging benchmarks}} \\
CT-RATE~\citep{hamamci2024ctrate} & 2024 & 21k pts / 25.7k vols & \bvnone & \bvfull & \bvnone & \bvfull & CT report gen \\
M3D-Bench~\citep{bai2024m3d} & 2024 & 120k pairs / 662k instr & \bvnone & \bvfull & \bvnone & \bvfull & 3D CT VQA \\
RadFM / RadBench~\citep{wu2025radfm} & 2023--25 & 13M 2D + 615k 3D scans & \bvnone & \bvpartial & \bvfull & \bvfull & Web 2D+3D VQA \\
3D-RAD~\citep{tang20253drad} & 2025 & 136k train / 170k QA & \bvfull & \bvfull & \bvnone & \bvpartial & Long.\ 3D CT VQA \\
Decipher-MR~\citep{ge2025deciphermr} & 2025 & 22k studies / 200k series & \bvnone & \bvfull & \bvfull & \bvfull & 3D MRI Dx/retrieval \\
\midrule
\multicolumn{8}{l}{\emph{Brain-MRI specialised}} \\
RadGenome-Brain / AutoRG-Brain~\citep{lei2024radgenomebrain} & 2024 & 1{,}007 cases / 3{,}408 pairs & \bvnone & \bvfull & \bvfull & \bvfull & Grounded brain report gen \\
BrainMD / BrainBench~\citep{wang2024brainmd} & 2024 & 2{,}453 scans & \bvnone & \bvpartial & \bvpartial & \bvfull & Slice-selected VQA \\
MM-NeuroOnco~\citep{guo2026mmneuroonco} & 2026 & 24.7k slices / 200k instr & \bvnone & \bvnone & \bvfull & \bvnone & 2D tumor instr.-tune \\
OmniBrainBench~\citep{peng2025omnibrainbench} & 2025 & 9{,}527 VQA / 31.7k imgs & \bvnone & \bvnone & \bvfull & \bvpartial & Tri-plane 2D clinical VQA \\
NeuroVLM-Bench~\citep{trojachanec2026neurovlmbench} & 2026 & 20 MLLMs evaluated & \bvnone & \bvnone & \bvfull & \bvnone & 2D clinical reasoning \\
BraTS (2012--24)~\citep{menze2015brats} & 2012--24 & \textasciitilde{}4{,}500 cases (2024) & \bvpartial & \bvfull & \bvfull & \bvnone & Tumor segmentation \\
IXI~\citep{ixiDataset} & $\sim$2006 & \textasciitilde{}600 subjects & \bvnone & \bvfull & \bvfull & \bvnone & Analysis / synthesis \\
\midrule
\multicolumn{8}{l}{\emph{Longitudinal biomarker challenges}} \\
OASIS-3~\citep{lamontagne2019oasis3} & 2019 & 1{,}378 pts / 2{,}842 sess & \bvfull & \bvfull & \bvfull & \bvnone & AD progression \\
ADNI / TADPOLE~\citep{marinescu2020tadpole} & 2017--19 & 1{,}677 pts & \bvfull & \bvfull & \bvfull & \bvnone & Biomarker forecast \\
\midrule
\textbf{BrainTRACE (ours)} & \textbf{2026} & \textbf{1{,}778 pts / 7{,}299 studies} & \bvfull & \bvfull & \bvfull & \bvfull & \textbf{5-level / 4-property suite} \\
\bottomrule
\end{tabular}
}
\end{table}

%% file: sections/B_appendix_template_tables.tex

\section{Dataset construction and task coverage}
\label{app:template-details}

This appendix chapter documents how the retained MR-RATE follow-up
cohort is converted into auditable evaluation instances. The emphasis is
on provenance, release boundaries, leakage control, and task coverage:
which records are model-facing, which records define hidden
report-grounded references, and which clinical reasoning depth and evidence
attributes each task family is intended to test.

\subsection{Construction and audit records}
\label{app:construction-artifacts}

The construction record has two purposes: it defines what an evaluated
model may see, and it preserves the evidence needed to audit the
reference after scoring. For item $i$, we write the model-facing record
as
\[
X_i=\{q_i, a_i, m_i, v_i\},
\]
where $q_i$ is the rendered question, $a_i$ is the answer format and any
visible options, $m_i$ is the permitted metadata, and $v_i$ is the
rendered visual evidence. The non-model-facing reference record is
\[
R_i=\{y_i, r_i, e_i, c_i\},
\]
where $y_i$ is the target answer or reference response, $r_i$ is the
rubric, $e_i$ is the report- or measurement-supported evidence, and
$c_i$ records exclusion, redaction, and scoring checks. Only $X_i$ is
shown during model evaluation. $R_i$ is retained for reconstruction,
leakage review, factual audit, and reproducibility.

\paragraph{\emph{1. Clinical eligibility and task specification.}}
Each task family first fixes the clinical reasoning depth to be tested, the
required evidence attributes, the answer format, and the allowed
metadata. Eligibility rules then determine whether a retained patient
and study sequence can support that question. For example, an
interval-change item requires comparable findings across timepoints,
whereas a volumetric item requires a segmentation-derived measurement
linked to the relevant study.

\paragraph{\emph{2. Reference formation.}}
Eligible report statements and segmentation-derived measurements are
converted into a hidden reference. Closed-form items store a normalized
target answer. Open-ended items store required facts, critical rubric
slots, accepted paraphrases, and clinically material contradictions.
Case-reasoning items store the same information separately for each of
the six decomposed steps, so a failure can be assigned to a specific
part of the longitudinal reasoning sequence rather than only to the
final case answer.

\paragraph{\emph{3. Model-facing instance construction.}}
The released question record contains the text seen by the model, the
answer format, any answer options, the permitted demographic or
timepoint metadata, and the visual-evidence specification. It does not
include the source report, evidence quotations, segmentation masks,
measurement tables, or the hidden reference. Broad-compatible models
receive rendered views; 3D-evidence items additionally preserve the
volume pathway needed by native-volume baselines.

\paragraph{\emph{4. Leakage and evidence-availability review.}}
Before release, each item is checked for redaction, report support,
visual-evidence availability, and consistency between the visible
question and the hidden reference. Closed-form items also receive a
text-only leakage probe using a stripped version of the question and
metadata. Items that fail rendering or reference checks are listed in
the exclusion records rather than removed without trace.

\paragraph{\emph{5. Scoring and reproducibility.}}
During evaluation, model outputs are joined back to the hidden reference
only at scoring time. Closed-form answers are scored by deterministic
normalization and exact match. Open-ended and case-level answers are
graded against the factual-audit rubric described in
Appendix~\ref{app:open-scoring-protocol}. The companion
repository contains the corresponding scoring specification, rendering
instructions, evaluation lists, exclusion records, and scoring
implementation:
\url{https://github.com/lanqz7766/BrainTRACE}. These
materials define the released denominator and allow an external user to
reconstruct the benchmark from the same MR-RATE access conditions.

\subsection{Upstream MR-RATE source and retained follow-up cohort}
\label{app:cohort-composition}

BrainTRACE is constructed from MR-RATE~\citep{mrrate2026}, a public
MRI corpus with paired clinical radiology reports. The upstream corpus
contains 83{,}425 patients, 98{,}334 MRI studies, and 705{,}254 MRI
series or metadata rows. BrainTRACE retains the longitudinal subset
needed for follow-up reasoning: 1{,}778 patients, 7{,}299 MRI studies,
and 65{,}974 series-level metadata rows, with at least three MRI
studies per retained patient. This source design makes provenance
auditable and keeps the references grounded in authentic, anonymized
radiology reports rather than generated report text. It does not imply
population representativeness for all neuroradiology indications.
BrainTRACE does not redistribute raw imaging or model-facing reports;
users obtain upstream MR-RATE access under its terms, and evaluated
models receive only the visual evidence permitted by each task.

\begin{figure}[h]
\centering
\includegraphics[width=0.70\linewidth]{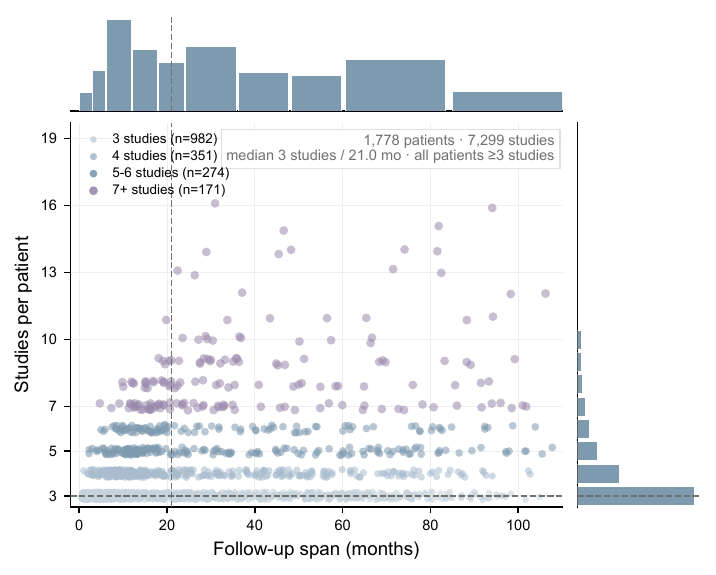}
\caption{The retained MR-RATE subset provides the longitudinal follow-up
structure required by BrainTRACE. Every retained patient has at least
three MRI studies, allowing report-grounded questions to be tied to
serial imaging rather than only single-visit findings.}
\label{fig:retained-cohort-distribution}
\end{figure}

\begin{figure}[h]
\centering
\includegraphics[width=0.92\linewidth]{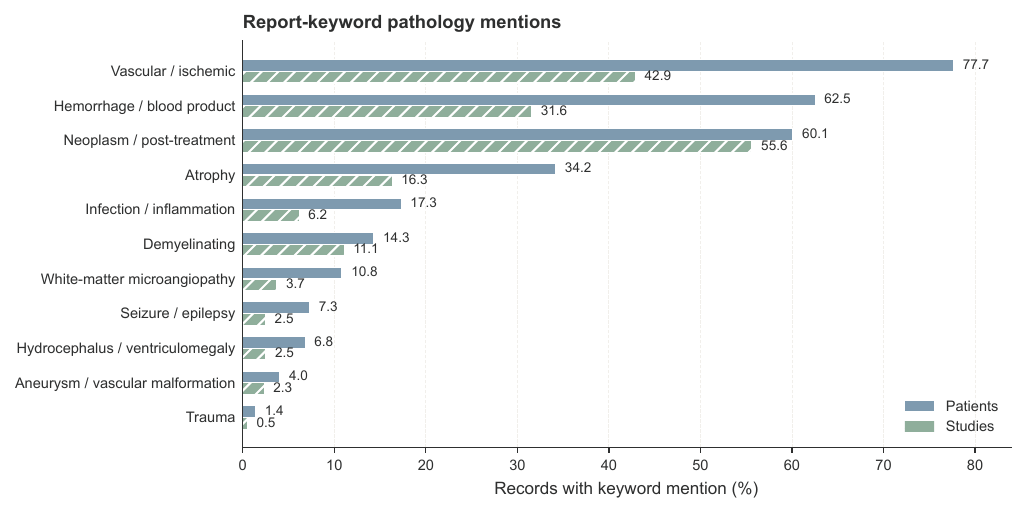}
\caption{Report-keyword audits indicate that the retained cohort is not
a single-pathology collection. Groups are non-mutually exclusive
report-text keyword audits used for cohort characterization, not
official diagnostic prevalence labels.}
\label{fig:pathology-keyword-distribution}
\end{figure}

\begin{figure}[h]
\centering
\includegraphics[width=\linewidth]{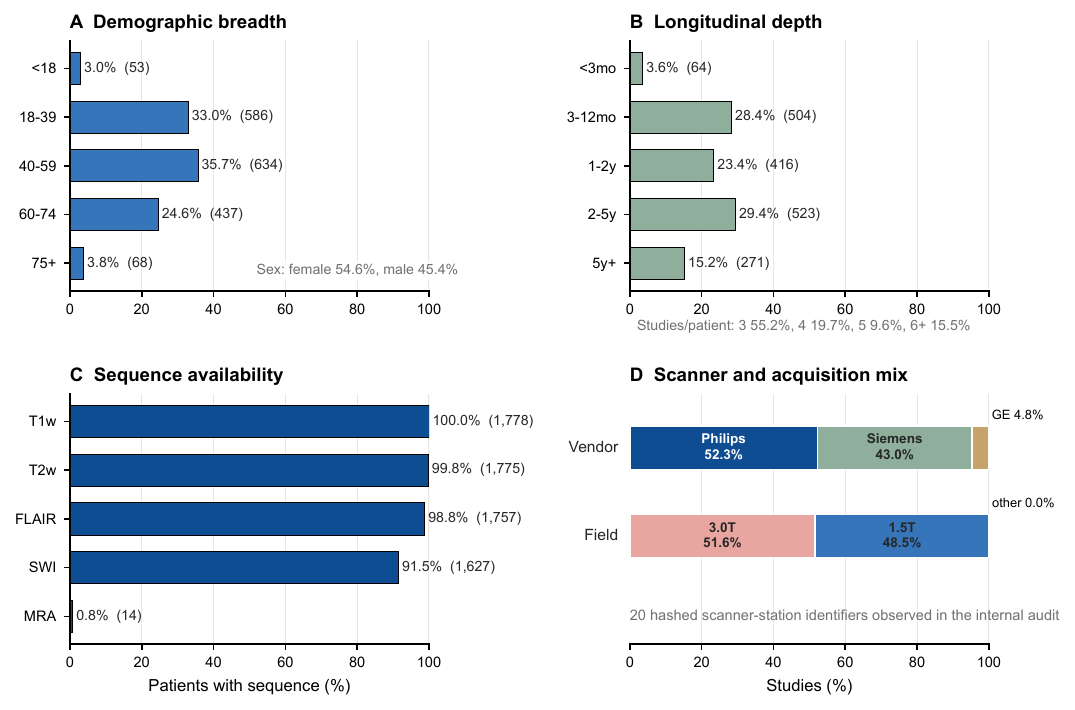}
\caption{The retained cohort is not restricted to a single demographic
or acquisition slice. Patient-level demographics and follow-up depth
show broad age coverage, balanced sex distribution, and substantial
multi-year follow-up. Sequence availability is high for T1w, T2w,
FLAIR, and SWI, while study-level acquisition metadata spans multiple
scanner vendors and both 1.5T and 3.0T field strengths. Percentages are
patient-level except for vendor and field strength, which are
study-level; the figure describes retained-cohort breadth rather than
population prevalence.}
\label{fig:cohort-diversity-dashboard}
\end{figure}

\paragraph{Cohort diversity.}
The retained follow-up cohort is not restricted to a single demographic,
sequence protocol, or scanner condition. Figure~\ref{fig:cohort-diversity-dashboard}
summarizes this evidence: patients span pediatric through older-adult
age groups, nearly half have follow-up beyond two years, common MRI
sequences are widely available, and study-level acquisition metadata
includes multiple vendors and both 1.5T and 3.0T scans. Figure~\ref{fig:pathology-keyword-distribution} complements this view by auditing report keyword mentions across broad pathology groups, showing that the retained cohort is not dominated by a single report category. These summaries
support the breadth of the retained cohort for benchmark construction;
they are not intended as prevalence or population representativeness
claims.

\FloatBarrier
\subsection{Benchmark task coverage}
\label{app:benchmark-task-coverage}

\begin{figure}[h]
\centering
\includegraphics[width=0.78\linewidth]{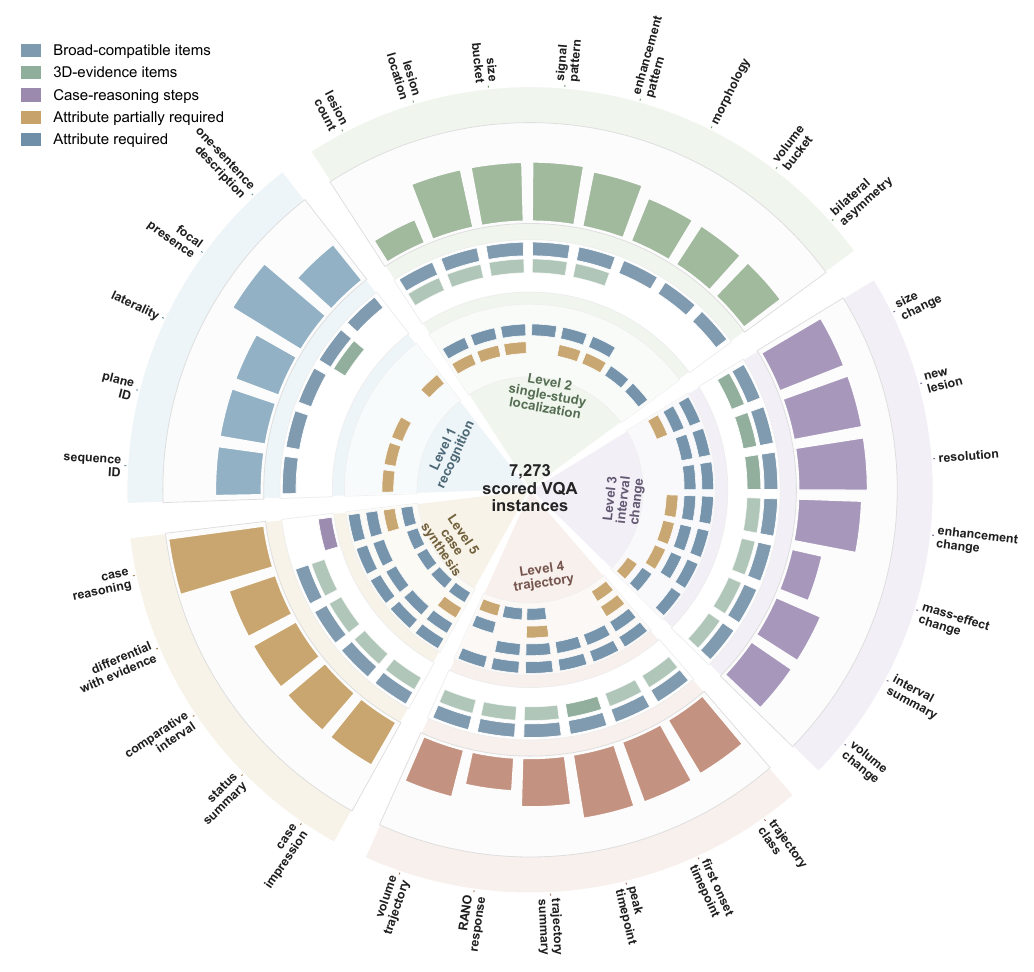}
\caption{Scored VQA instances span five clinical-operation levels,
three evidence conditions, and the evidence attributes used to audit
longitudinal, multi-sequence, volumetric, and clinically interpretive
reasoning.}
\label{fig:benchmark-coverage-circos}
\end{figure}

The retained source cohort is sampled into evaluated VQA instances that
span clinical reasoning depth, evidence conditions, and evidence attributes.
Figure~\ref{fig:benchmark-coverage-circos} gives the visual overview;
Table~\ref{tab:tmpl-props} records the evaluated denominators and the
intended evidence demands for each task family.

\FloatBarrier
\subsection{Clinical-level taxonomy}
\label{app:clinical-taxonomy}

BrainTRACE organizes questions by the clinical reasoning depth required to
answer them, rather than treating all items as interchangeable visual
recognition tasks. The levels therefore define what kind of evidence
operation is being tested; they are not merely ordered difficulty bins.

\begin{itemize}
    \item \textbf{L1: Image and acquisition recognition.} These
    questions test whether a model can identify acquisition properties
    and basic visual findings, including sequence, plane, laterality,
    focal abnormality, and concise image description.

    \item \textbf{L2: Anatomical localization and single-study finding
    recognition.} These questions require lesion counting, anatomical
    localization, size or volume categorization, signal and enhancement
    characterization, morphology assessment, and asymmetry recognition
    within a single study.

    \item \textbf{L3: Disease-specific interpretation and interval
    change.} These questions compare findings across studies, including
    size change, new lesion appearance, resolution, enhancement
    evolution, mass-effect change, interval summaries, and volume
    change.

    \item \textbf{L4: Longitudinal trajectory or quantitative change
    assessment.} These questions require reasoning over trajectories,
    including progression pattern, first onset, peak timepoint,
    narrative trajectory, response category, and serial volume change.

    \item \textbf{L5: Case-level and integrated clinical judgment.} These questions
    require synthesis across visual evidence and clinical context,
    including impression generation, disease-status assessment,
    comparative summaries, and differential diagnosis with supporting
    evidence.
\end{itemize}

Per-track denominators and evidence-property annotations are reported in
Table~\ref{tab:tmpl-props}.

\subsection{Per-task counts and evidence-property coverage}
\label{app:template-counts}
\label{app:template-properties}

Table~\ref{tab:tmpl-props} combines task-family sampling provenance
with the evidence attributes used throughout the paper. The count
columns define evaluated denominators for the broad-compatible,
3D-evidence, and case-reasoning tracks. The rightmost columns state the
intended evidence demands, which makes the table a task design matrix
rather than a second prose taxonomy. We split volumetric evidence into
\emph{V}$_{\!p}$ for pseudo-3D multi-plane or multi-slice composites and
\emph{V}$_{\!n}$ for native-volume segmentation-derived values.

\begin{table}[h]
\centering
\scriptsize
\setlength{\tabcolsep}{2.2pt}
\caption{Task-family design matrix defines evaluated denominators and
intended evidence demands. Broad, 3D, and Case give evaluated
VQA-instance counts. \textbf{L} = longitudinal, \textbf{M} =
multi-sequence, \textbf{V}$_{\!p}$ = pseudo-3D volumetric,
\textbf{V}$_{\!n}$ = native-volume volumetric, and \textbf{C} =
clinical reasoning beyond surface perception.
\protect\bvfull{} = required, \protect\bvpartial{} = partial or
compatible, and \protect\bvnone{} = not required; \protect\bvfull{}$^\ast$
marks a central purpose.}
\label{tab:tmpl-counts}
\label{tab:tmpl-props}
\resizebox{\linewidth}{!}{%
\begin{tabular}{llrrrrlccccc}
\toprule
Task family & Ans. & Broad & 3D & Case & Total & Notes & \textbf{L} & \textbf{M} & \textbf{V}$_{\!p}$ & \textbf{V}$_{\!n}$ & \textbf{C} \\
\midrule
L1.1 pulse\_sequence\_identification & MCQ & 200 & 0 & 0 & 200 & & \bvnone & \bvnone & \bvpartial & \bvnone & \bvnone \\
L1.2 plane\_id & MCQ & 200 & 0 & 0 & 200 & & \bvnone & \bvnone & \bvpartial & \bvnone & \bvnone \\
L1.3 laterality\_of\_focal\_finding & MCQ & 200 & 0 & 0 & 200 & & \bvnone & \bvnone & \bvpartial & \bvnone & \bvnone \\
L1.4 focal\_abnormality\_presence & binary & 200 & 150 & 0 & 350 & & \bvnone & \bvnone & \bvnone & \bvnone & \bvnone \\
L1.6 one\_sentence\_image\_description & open & 200 & 0 & 0 & 200 & & \bvnone & \bvnone & \bvpartial & \bvnone & \bvnone \\
\midrule
L2.1 discrete\_abnormality\_count & ord. & 85 & 24 & 0 & 109 & pool-limited & \bvnone & \bvfull & \bvpartial & \bvnone & \bvnone \\
L2.2 location\_bucket\_dominant & MCQ & 200 & 57 & 0 & 257 & canonical n & \bvnone & \bvfull & \bvpartial & \bvnone & \bvnone \\
L2.3 size\_bucket\_dominant & ord. & 200 & 57 & 0 & 257 & canonical n & \bvnone & \bvfull & \bvpartial & \bvnone & \bvnone \\
L2.4 signal\_pattern & MCQ & 200 & 58 & 0 & 258 & & \bvnone & \bvfull & \bvnone & \bvnone & \bvnone \\
L2.5 enhancement\_pattern & MCQ & 188 & 54 & 0 & 242 & dedup & \bvnone & \bvfull & \bvpartial & \bvnone & \bvnone \\
L2.6 morphology\_and\_boundary & open & 200 & 0 & 0 & 200 & & \bvnone & \bvfull & \bvpartial & \bvnone & \bvnone \\
L2.7 structure\_volume\_bucket & ord. & 200 & 0 & 0 & 200 & & \bvnone & \bvnone & \bvnone & \bvfull$^\ast$ & \bvnone \\
L2.8 bilateral\_volumetric\_asymmetry & MCQ & 200 & 0 & 0 & 200 & & \bvnone & \bvnone & \bvnone & \bvfull$^\ast$ & \bvnone \\
\midrule
L3.1 size\_change\_at\_site & ord. & 200 & 100 & 0 & 300 & & \bvfull & \bvfull & \bvpartial & \bvnone & \bvnone \\
L3.2 new\_focal\_abnormality & binary & 200 & 101 & 0 & 301 & & \bvfull & \bvfull & \bvnone & \bvnone & \bvnone \\
L3.3 resolution\_of\_prior & MCQ & 200 & 100 & 0 & 300 & & \bvfull & \bvfull & \bvnone & \bvnone & \bvnone \\
L3.4 enhancement\_change\_at\_site & MCQ & 200 & 75 & 0 & 275 & & \bvfull & \bvfull & \bvpartial & \bvnone & \bvnone \\
L3.5 mass\_effect\_change\_at\_site & MCQ & 88 & 45 & 0 & 133 & pool-limited & \bvfull & \bvfull & \bvpartial & \bvnone & \bvnone \\
L3.6 open\_interval\_change\_at\_site & open & 200 & 11 & 0 & 211 & & \bvfull & \bvfull & \bvpartial & \bvnone & \bvnone \\
L3.7 structure\_volume\_change & ord. & 200 & 11 & 0 & 211 & canonical n & \bvfull & \bvnone & \bvnone & \bvfull$^\ast$ & \bvpartial \\
\midrule
L4.1 trajectory\_class\_closed & MCQ & 195 & 77 & 0 & 272 & & \bvfull & \bvfull & \bvpartial & \bvnone & \bvpartial \\
L4.2 new\_lesion\_timing & ord. & 200 & 78 & 0 & 278 & & \bvfull & \bvfull & \bvnone & \bvnone & \bvnone \\
L4.3 peak\_extent\_timepoint & ord. & 200 & 80 & 0 & 280 & & \bvfull & \bvfull & \bvnone & \bvnone & \bvnone \\
L4.4 open\_trajectory\_summary & open & 200 & 11 & 0 & 211 & & \bvfull & \bvfull & \bvpartial & \bvnone & \bvfull \\
L4.5 response\_proxy\_RANO\_lite & MCQ & 97 & 46 & 0 & 143 & pool-limited & \bvfull & \bvfull & \bvnone & \bvnone & \bvfull$^\ast$ \\
L4.6 structure\_volume\_trajectory & ord. & 200 & 10 & 0 & 210 & canonical n & \bvfull & \bvnone & \bvnone & \bvfull$^\ast$ & \bvpartial \\
\midrule
L5.1 full\_impression\_3to5\_sentences & open & 200 & 14 & 0 & 214 & canonical n & \bvfull & \bvfull & \bvpartial & \bvnone & \bvfull$^\ast$ \\
L5.2 brief\_imaging\_status\_summary & open & 200 & 14 & 0 & 214 & canonical n & \bvfull & \bvfull & \bvnone & \bvnone & \bvfull \\
L5.3 comparative\_interval\_summary & open & 200 & 14 & 0 & 214 & canonical n & \bvfull & \bvfull & \bvnone & \bvnone & \bvfull \\
L5.4 differential\_with\_visual\_evidence & open & 200 & 13 & 0 & 213 & & \bvfull & \bvfull & \bvnone & \bvnone & \bvfull$^\ast$ \\
Decomposed case reasoning & case & 0 & 0 & 420 & 420 & case-level & \bvfull & \bvfull & \bvnone & \bvnone & \bvfull$^\ast$ \\
\midrule
\textbf{Total} & & \textbf{5{,}653} & \textbf{1{,}200} & \textbf{420} & \textbf{7{,}273} & & & & & & \\
\bottomrule
\end{tabular}
}
\end{table}

\section{Scoring, audit, and reproducibility protocol}
\label{app:scoring-audit}

This appendix chapter defines the factual-audit rule used for
open-ended answers and decomposed case-reasoning steps. The goal is to
measure clinical acceptability under the released report-grounded
rubric, not stylistic similarity to a reference sentence. The chapter
also states how Pass \%, Step Pass, and Case Success are computed and
reports agreement and judge-family sensitivity audits of the scoring
instrument.

\subsection{Factual-audit scoring for open-ended and case-level answers}
\label{app:open-scoring-protocol}

Open-ended and case-level answers are scored by factual clinical audit
rather than lexical similarity. A response passes only if it covers the
required clinical facts, satisfies the critical rubric slots, avoids
clinically material contradictions, and does not introduce unsupported
claims that would alter the interpretation. This distinction is
necessary for radiology-style answers: fluent prose can still fail if it
changes laterality, reverses interval direction, assigns a finding to
the wrong anatomy, cites the wrong timepoint, or asserts an unsupported
disease-status conclusion.

For each open-ended item or decomposed case step $i$, the evaluator
receives the question $q_i$, the model answer $y_i$, the reference
answer $r_i$, accepted paraphrases when available, required clinical
facts, critical rubric slots, and pre-specified failure criteria. The
evaluator does not receive the source report, segmentation tables,
evidence quotations, or construction record. All evaluated outputs use
the same judge model version, instruction text, response schema, and
deterministic decoding setting, and the complete judge record is
archived for each item.

\subsubsection{Scoring fields and pass rule}
\label{app:scoring-fields}

For each item or decomposed case step, the judge record contains the
following fields. These fields define the scoring variables used in the
reported metrics:
\begin{itemize}
\item \texttt{required\_fact\_results}: checks whether named findings,
locations, laterality, timepoints, trajectory direction, and current
status required by the reference are explicitly present or clinically
equivalent.
\item \texttt{slot\_results}: evaluates each rubric slot against the
model answer. Critical slots enter the pass rule; noncritical slots
are retained for coverage and quality analyses.
\item \texttt{critical\_errors}: records hard contradictions such as
wrong anatomy, wrong laterality, wrong interval direction, wrong
timepoint, wrong modality-specific reading, wrong diagnosis, or
clinically material severity errors.
\item \texttt{unsupported\_claims}: records claims not supported by the
reference or rubric, such as a new lesion, disease progression,
treatment response, or diagnosis absent from the clinical reference.
\item \texttt{clinically\_acceptable}: gives the final rubric-based
factual verdict used for Pass \%.
\item \texttt{quality\_score}: records graded answer quality on the
same 1--5 clinical scale used in the human validation subset.
\end{itemize}

\paragraph{Pass rule.}
Let $F_i=1$ if all required facts for item $i$ are present or
clinically equivalent, $S_i=1$ if all critical rubric slots are
satisfied, $E_i=1$ if no critical error is present, and $U_i=1$ if no
clinically material unsupported claim is present. The item-level pass
indicator is
\[
P_i = F_i \cdot S_i \cdot E_i \cdot U_i .
\]
Equivalently, an answer fails if it omits a required fact, misses a
critical rubric slot, contradicts the reference, or introduces a
clinically material unsupported claim.

\paragraph{Reported metrics.}
For a set $\mathcal{I}$ of open-ended items, Pass \% is
\[
\mathrm{Pass}(\mathcal{I}) =
100 \times \frac{1}{|\mathcal{I}|}\sum_{i\in\mathcal{I}} P_i .
\]
For the case-reasoning track, each case $c$ has six ordered steps
$s=1,\ldots,6$. With $P_{c,s}$ denoting the pass indicator for step
$s$ in case $c$, Step Pass and Case Success are
\[
\mathrm{StepPass} =
100 \times \frac{1}{6|\mathcal{C}|}
\sum_{c\in\mathcal{C}}\sum_{s=1}^{6} P_{c,s},
\qquad
\mathrm{CaseSuccess} =
100 \times \frac{1}{|\mathcal{C}|}
\sum_{c\in\mathcal{C}}\prod_{s=1}^{6} P_{c,s}.
\]
Case Success is stricter than Step Pass because all six steps must pass
within the same longitudinal case. We also report Mean Quality Score,
Critical Error \%, required-fact coverage, missing-required-fact
counts, slot-match rates, unsupported-claim counts, and the full judge
records. These quantities are diagnostic; the primary open-ended
leaderboard is ranked by Pass \%.

Let $Q_i \in \{1,\ldots,5\}$ denote the graded quality score, and let
$M_i$ and $K_i$ denote the number of matched and required facts for item
$i$. The diagnostic aggregates are
\[
\mathrm{MeanQS}(\mathcal{I}) =
\frac{1}{|\mathcal{I}|}\sum_{i\in\mathcal{I}} Q_i,\qquad
\mathrm{CriticalError}(\mathcal{I}) =
100 \times \frac{1}{|\mathcal{I}|}\sum_{i\in\mathcal{I}} (1-E_i),
\]
\[
\mathrm{RequiredFactCoverage}(\mathcal{I}) =
100 \times
\frac{\sum_{i\in\mathcal{I}} M_i}{\sum_{i\in\mathcal{I}} K_i}.
\]
These aggregates explain where failures arise, but they do not override
the binary factual-audit pass rule.

The audit uses a conservative interpretation of clinical acceptability
under the released rubric. Generic radiology language does not satisfy a
named finding, laterality, trajectory, or disease-status requirement.
Direction reversals, laterality errors, wrong-anatomy claims,
wrong-diagnosis claims, and unsupported new findings are failures even
when the surrounding answer is fluent. Omissions and contradictions are
recorded separately: an omission affects required-fact coverage, while a
contradiction is counted as a critical error.

Human validation is reported in Appendix~\ref{appx:iaa}. Two expert
reviewers independently scored a case-reasoning subset under the same
clinical quality and critical-error semantics, and the released records
include per-step expert grades and judge fields so that external
auditors can repeat or extend the agreement analysis.
Appendix~\ref{app:case-examples} provides annotated success and failure
examples showing how the audit treats matched required facts, missing
facts, wrong direction, wrong anatomy, and unsupported clinical claims.

\subsection{Open-ended audit leaderboards}
\label{app:open-leaderboards}

Table~\ref{tab:bulk-open-v06-full} gives the complete open-ended
leaderboard behind the abbreviated main-text summary. It is retained in
the appendix so the main paper can emphasize scoring design and
headline patterns while still making every evaluated configuration
auditable. The table reports the factual-audit metrics available for
the completed broad-compatible open-ended runs, together with four
traditional text-overlap or embedding-similarity diagnostics. These
secondary diagnostics are included for comparability with prior VQA and
report-generation evaluations, but the table remains ranked by factual
audit because BLEU~\citep{papineni2002bleu},
ROUGE~\citep{lin2004rouge}, chrF++~\citep{popovic2017chrfpp}, and
BERTScore~\citep{zhang2020bertscore} do not check required clinical
slots, contradictions, laterality, timing, or unsupported disease-status
claims.

\begin{table}[h]
\centering
\scriptsize
\setlength{\tabcolsep}{3pt}
\caption{The full broad-compatible open-ended leaderboard is ranked by
factual-audit Pass \%, while traditional text metrics are reported only
as secondary comparability diagnostics. QS is Mean Quality Score and
Crit. is Critical Error \%; BLEU4, ROUGE-L, chrF++, and BERTScore are
mean scores over the same broadQA open-ended items in their native
scales. Bold marks the best value in each metric column and underlining
marks the second-best value; lower is better for Crit., and higher is
better for all other metric columns.}
\label{tab:bulk-open-v06-full}
\resizebox{\linewidth}{!}{%
\begin{tabular}{r l r r r r r r r}
\toprule
\textbf{Rank} & \textbf{Model} & \textbf{Pass} & \textbf{QS} & \textbf{Crit.} & \textbf{BLEU4} & \textbf{ROUGE-L} & \textbf{chrF++} & \textbf{BERTScore} \\
\midrule
1 & GPT-5.4 & \textbf{19.8} & \textbf{2.55} & 22.2 & 0.065 & 0.198 & \underline{0.281} & 0.043 \\
2 & Qwen3-VL-30B & \underline{14.4} & 2.27 & 36.6 & 0.039 & 0.193 & 0.273 & 0.149 \\
3 & Gemini 2.5 Pro & 14.2 & 2.28 & 32.9 & 0.040 & 0.183 & 0.267 & 0.144 \\
4 & Qwen3-VL-8B & 13.8 & \underline{2.29} & 29.0 & \textbf{0.081} & 0.202 & 0.272 & 0.157 \\
5 & Qwen3-VL-4B & 11.8 & 2.24 & 28.0 & 0.070 & 0.190 & 0.263 & 0.134 \\
6 & GPT-5-mini & 10.6 & 2.07 & 40.3 & 0.056 & 0.176 & 0.264 & 0.102 \\
7 & GPT-5 & 10.5 & 2.12 & 33.8 & \underline{0.074} & 0.179 & 0.263 & 0.128 \\
8 & HuatuoGPT-Vision-7B & 9.9 & 2.13 & 25.6 & 0.072 & \underline{0.208} & \textbf{0.289} & 0.157 \\
9 & Claude Opus 4.6 & 9.6 & 2.00 & 43.0 & 0.020 & 0.142 & 0.247 & 0.050 \\
10 & InternVL3-14B & 8.6 & 2.11 & 32.2 & 0.044 & 0.206 & 0.276 & \underline{0.181} \\
11 & Qwen3.5-27B & 8.1 & 1.91 & 43.7 & 0.052 & 0.148 & 0.211 & -1.116 \\
12 & InternVL3-78B & 7.6 & 2.04 & 34.6 & 0.037 & 0.199 & 0.276 & 0.161 \\
13 & Lingshu-7B & 6.8 & 2.03 & 25.7 & 0.042 & \textbf{0.211} & 0.273 & \textbf{0.185} \\
14 & InternVL3-8B & 6.6 & 2.00 & 32.1 & 0.045 & 0.207 & 0.272 & 0.166 \\
15 & MedGemma-4B-it & 5.4 & 1.90 & 33.2 & 0.026 & 0.176 & 0.239 & 0.141 \\
16 & Qwen2.5-VL-7B & 4.9 & 1.96 & 27.4 & 0.037 & 0.204 & 0.251 & 0.164 \\
17 & RadFM & 4.3 & 1.96 & \underline{21.4} & 0.014 & 0.108 & 0.195 & -0.107 \\
18 & LLaVA-Med-v1.5 & 3.8 & 2.06 & \textbf{8.6} & 0.010 & 0.070 & 0.105 & -1.605 \\
19 & Janus-Pro-7B & 1.8 & 1.86 & 24.4 & 0.007 & 0.082 & 0.180 & -0.529 \\
\bottomrule
\end{tabular}
}
\end{table}

Pass \%, QS, and Critical Error \% are rubric-based factual-audit
metrics: they score clinical acceptability under the released reference,
graded answer quality, and disqualifying clinical errors. The lexical
and embedding metrics are deliberately secondary. A high-overlap answer
can still omit a required fact or add an unsupported diagnosis, while an
acceptable answer can use wording that differs from the reference.
BERTScore uses the rescaled RoBERTa-large F1 convention in the scored
artifacts, so negative values are possible.

\subsection{Case-level reasoning audit records}
\label{app:case-reasoning-audit}

Table~\ref{tab:open-v06} reports the headline case-level columns in the
main paper, so this appendix focuses on the released audit records
rather than duplicating a case-only leaderboard. Figure~\ref{fig:bootstrap-ci-forest}
summarizes Step Pass uncertainty for the strongest case-level
configurations. The release retains the per-step and per-case records
needed to recompute Step Pass, Case Success, Mean QS, Critical Error,
missing-required-fact counts, and required-fact coverage diagnostics.

\subsection{Per-step attrition in case-reasoning}
\label{app:case-step-attrition}

Figure~\ref{fig:case-attrition} gives the per-step diagnostic:
Panel~A tracks cumulative case survival, while Panel~B shows which
decomposed reasoning steps pass before a case collapses. The main
pattern is not a single weak model row but a consistent composition
failure. Earlier steps can pass at roughly 30--43\% for the strongest
configurations, but final synthesis remains near zero. Claude Opus 4.6
has the highest final-synthesis pass rate among the displayed rows
(7.1\%), but still has 0/70 full-case survival; GPT-5-mini has the
highest aggregate Step Pass (24.5\%) but also reaches 0/70 full-case
survival. Thus Step Pass is useful for localizing failures, whereas
Case Success tests whether those local successes occur in the same
longitudinal case.

The GPT-5-family rows remain closely spaced in aggregate Step Pass, as
shown by the consolidated bootstrap intervals in
Figure~\ref{fig:bootstrap-ci-forest}. Increasing GPT-5.4 reasoning
effort redistributes successes across step types but does not improve
aggregate Step Pass, reinforcing that the bottleneck is evidence
propagation and final synthesis rather than a single isolated
recognition step.

\subsection{Two-reviewer human validation of the factual-audit judge}
\label{appx:iaa}

\subsubsection{Protocol}
This analysis evaluates agreement between the released factual-audit
judge and expert review; it is not a clinical certification study. Two
board-certified neuroradiologists (Reviewer~1 and Reviewer~2)
independently graded the same 60-step subset, blind to each other and to
the LLM-judge output.
The subset comprises 10 longitudinal tumor cases $\times$ 6 reasoning
steps, with model outputs split equally between GPT-5.4 (5 cases) and
Qwen3-VL-30B (5 cases). Each step was rated on a 5-point clinical
quality scale (1 = critical error; 5 = clinically excellent) and a
binary critical-error flag. After an initial round, Reviewer~1 rescored
the subset under finalized rubric semantics; we use the post-calibration
grades. Total per-reviewer effort was approximately one hour.

\subsubsection{Headline metrics}
Quadratic-weighted Cohen's $\kappa_w$ on the 1--5 quality:
0.737 (Reviewer~1 vs.\ judge, 95\% CI [0.56, 0.86]),
0.645 (Reviewer~2 vs.\ judge, [0.47, 0.78]),
0.821 (Reviewer~1 vs.\ Reviewer~2, [0.72, 0.89]).
Intraclass correlation ICC(3,1) is 0.740 / 0.649 / 0.824 respectively.
Binary critical-error Cohen's $\kappa$:
0.475 (Reviewer~1 vs.\ judge, [0.29, 0.67]),
0.552 (Reviewer~2 vs.\ judge, [0.35, 0.74]),
0.767 (Reviewer~1 vs.\ Reviewer~2, [0.60, 0.90]).
Judge specificity is $1.000$ against both reviewer references;
sensitivity is $0.484$ and $0.536$ respectively.

\begin{figure}[h]
\centering
\includegraphics[width=\linewidth]{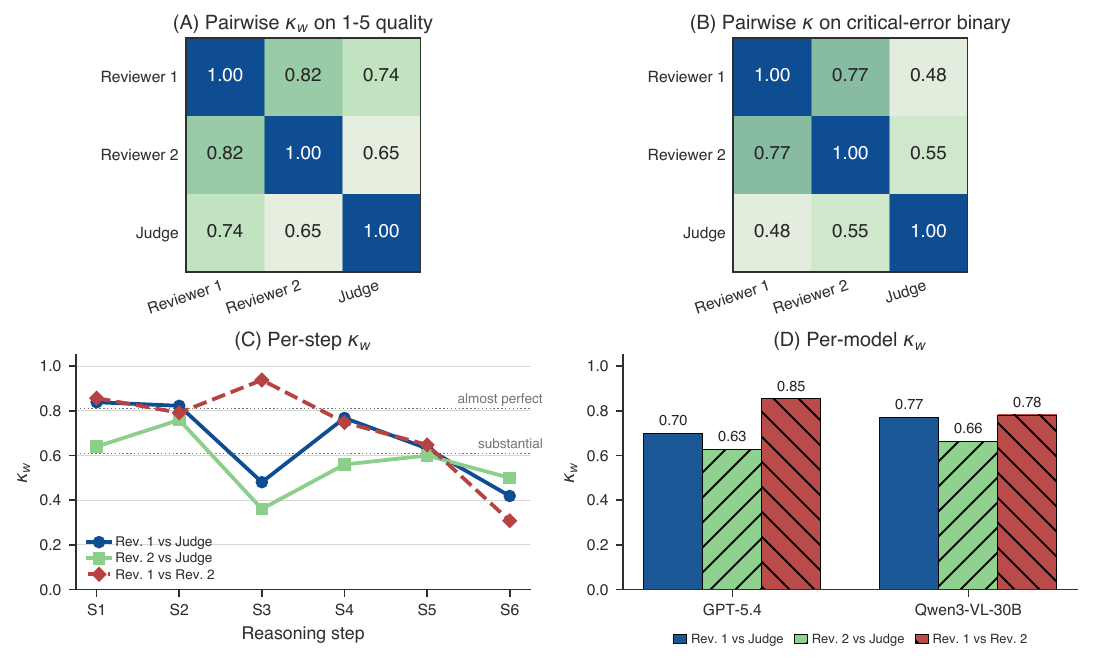}
\caption{Three-way inter-annotator agreement analysis evaluates the
factual-audit judge as a scoring instrument rather than treating it as
an unchecked automatic metric. The audit covers the case-reasoning track
($n=60$ decomposed VQA steps).
(A,B) Pairwise agreement on the 1--5 quality scale and the
critical-error binary.
(C) Per-step $\kappa_w$.
(D) Per-model $\kappa_w$.}
\label{fig:iaa}
\end{figure}

\subsubsection{Per-step decomposition}
Reviewer~1 vs.\ judge $\kappa_w$ across S1--S6:
0.84, 0.82, 0.48, 0.77, 0.63, 0.42.
Reviewer~2 vs.\ judge: 0.64, 0.76, 0.36, 0.56, 0.60, 0.50.
Inter-human (Reviewer~1 vs.\ Reviewer~2): 0.86, 0.79, 0.94, 0.75, 0.65, 0.31.
Final-synthesis agreement degrades for all three pairs, consistent with
the greater ambiguity and evidence-composition burden of integrative
reasoning.

\subsubsection{Per-model decomposition}
Reviewer~1 vs.\ judge $\kappa_w$ is 0.699 on GPT-5.4 decomposed VQA
steps ($n=30$) and 0.772 on Qwen3-VL-30B decomposed VQA steps
($n=30$). Reviewer~2 vs.\ judge is 0.628 and 0.664 respectively, while
inter-human agreement is 0.855 and 0.782. Cross-model agreement is
similar in rank and magnitude, indicating the judge comparison is not
driven by a single model family.

\subsubsection{Scope of the validation}
The validation is scoped to two reviewers and 60 decomposed VQA steps
from the case-reasoning track. We position the result as an
agreement audit of the scoring instrument, not as an independent de novo
radiologist adjudication of the full benchmark. The full per-step expert
grades and judge raw fields are released so that external auditors may
extend the validation to additional reviewers, items, or scoring
specification revisions.

\subsection{Cross-family judge sensitivity audit}
\label{appx:judge-family-sensitivity}

The primary open-ended evaluator in the main experiments is
\texttt{gpt-4o}. To examine whether this choice affects
reported open-ended pass rates, we ran a small cross-family sensitivity
analysis on stratified broad open-ended items. The audit uses 52 items
per cell, sampled evenly across L1.6, L2.6, L3.6, and L4.4
($13$ items per family), and compares release-scored judgments from the
GPT judge with Gemini~2.5 Pro and Claude Sonnet~4.6 on outputs from
GPT-5.4, Claude Opus~4.6, and Gemini~2.5 Pro. All pass rates and
quality scores in this subsection are recomputed with the released
scoring implementation rather than ad hoc parsing of judge records.

\begin{figure}[h]
\centering
\includegraphics[width=\linewidth]{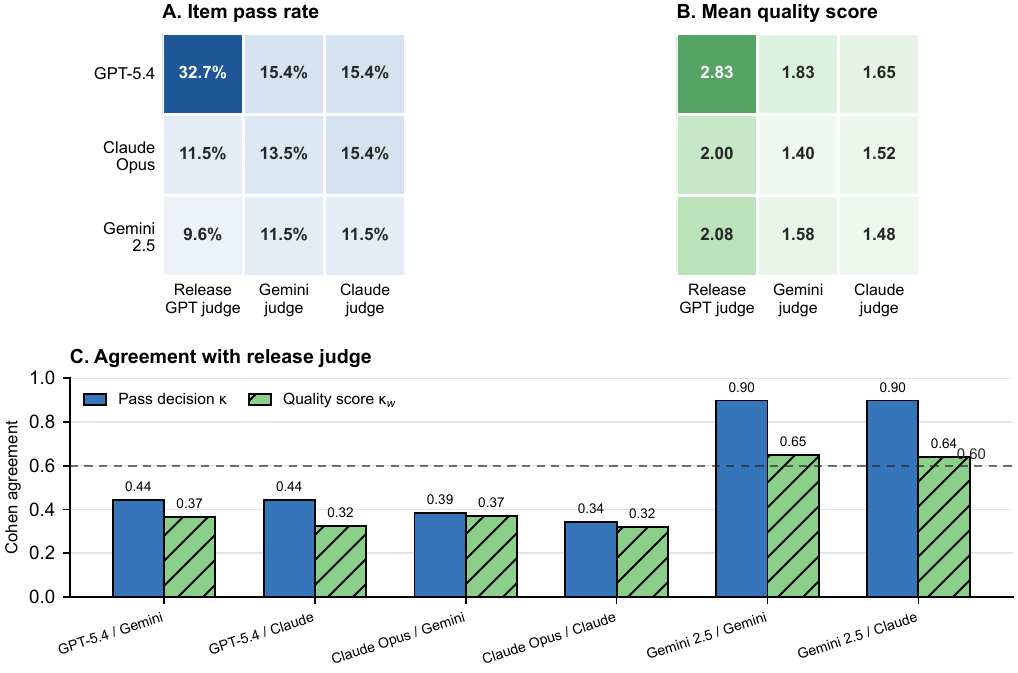}
\caption{Cross-family judge audit measures sensitivity to the judge
family under the released factual scoring rule. Each cell contains the
same stratified sample size ($n=52$). Alternative Gemini and Claude
judges preserve low absolute pass rates, while the agreement panel shows
where the release judge aligns most strongly with cross-family
alternatives.}
\label{fig:cross-family-judge-audit}
\end{figure}

The audit is intentionally narrow: it checks whether the released rubric
behaves differently when the judge family changes on a stratified
subset. The main sensitivity is localized to GPT-5.4 outputs. The
release judge passes 32.7\% of the sampled GPT-5.4 items, whereas both
cross-family judges pass 15.4\%; the discordance is mostly release-only
passes. For Claude Opus 4.6 and Gemini 2.5 Pro outputs, pass rates are
closer across judges, and Gemini-output agreement is high
($\kappa=0.898$ for pass decisions). We therefore treat exact
open-ended Pass \% values as rubric-and-judge measurements, while the
central empirical conclusion remains stable in this audit: clinically
acceptable open-ended answers are rare across judge families.

\subsection{Evaluation records and reproducibility}
\label{app:eval-provenance}

The public scoring specification is released as \texttt{docs/SCORING.md} in
the companion repository:
\url{https://github.com/lanqz7766/BrainTRACE/blob/main/docs/SCORING.md}. The
repository contains the rendering specifications, evaluation interfaces,
scoring implementation, canonical item lists, exclusion registries, and audit
records needed to recompute the reported scores. Closed-form scoring
uses exact matching after answer normalization and removal of visible
\texttt{<think>...</think>} text. Open-ended and case-level answers use
the factual-audit rule defined in
Appendix~\ref{app:open-scoring-protocol}.

For each track, the release records the evaluated item set, the model
answer, and the corresponding scoring record. Closed-form records
contain the canonical item identifier, the normalized answer, and the
exact-match decision. Open-ended records contain the rubric slots,
required-fact coverage, critical-error flag, unsupported-claim flag,
quality score, and judge identity. Case-reasoning records contain the
six ordered step decisions for each longitudinal case, which determine
Step Pass and Case Success under the formulas above. The release also
includes the exclusion registries used after rendering and reference
checks. These records define the denominators in the reported tables;
this paragraph documents reproducibility rather than adding another
scale statistic.

\begin{figure}[h]
\centering
\includegraphics[width=\linewidth]{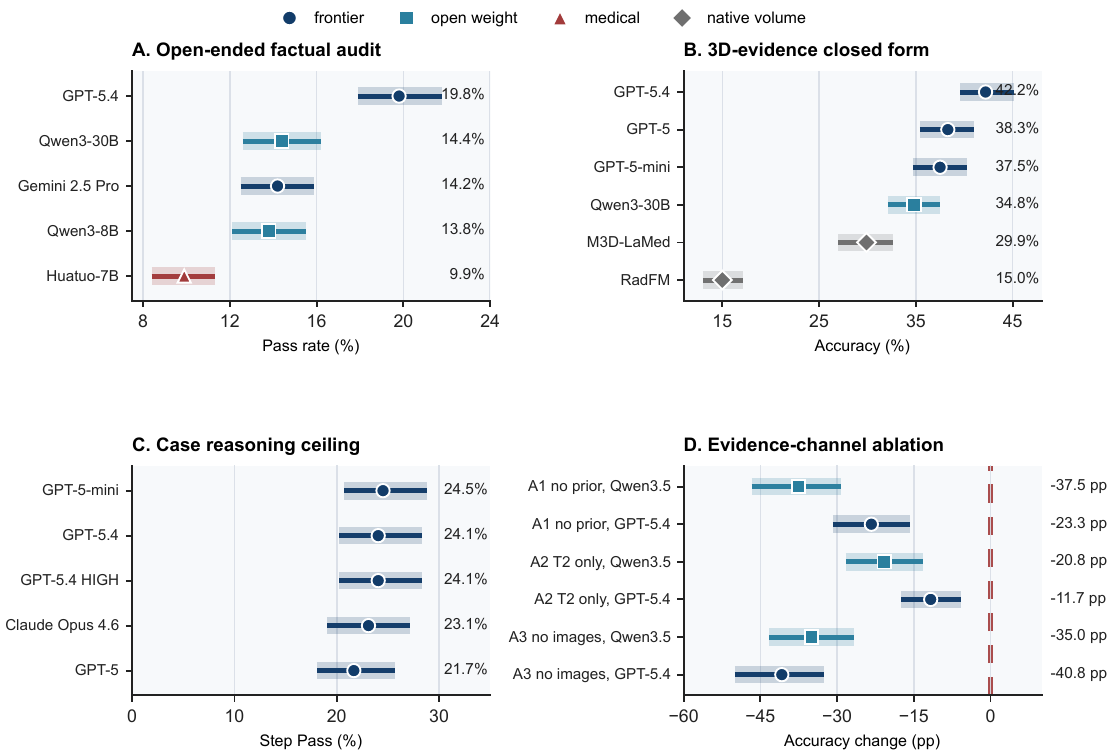}
\caption{Bootstrap confidence intervals summarize uncertainty in the
headline empirical pattern without adding a second leaderboard
narrative. Thick bands show 95\% bootstrap confidence intervals and
points show estimates. Panel D uses paired bootstrap intervals on
matched ablation samples; all intervals remain below zero. Intervals use
nonparametric percentile bootstrap over raw scored records or judge-step
records, matching the protocol summarized in
Appendix~\ref{app:eval-provenance}.}
\label{fig:bootstrap-ci-forest}
\end{figure}

Figure~\ref{fig:bootstrap-ci-forest} consolidates uncertainty estimates
across headline metrics. It is a visual uncertainty summary and replaces
separate appendix-only CI tables for the case-reasoning and
evidence-channel ablation analyses.

\section{Additional empirical diagnostics}
\label{app:empirical-diagnostics}

This appendix chapter collects diagnostics that should not be interpreted
as a second leaderboard. Each analysis isolates a different reading of
the same benchmark: performance under available 3D input interfaces,
rank instability across evaluation axes, and difficulty after correcting
for level-specific chance floors.

\subsection{Paper-reported 3D-evidence results}
\label{app:threed-leaderboard}

Table~\ref{tab:threed-paper-reported} is the numeric companion to the
main-text 3D column and to the representative 3D interval panel in
Figure~\ref{fig:bootstrap-ci-forest}. It is retained for confidence
interval provenance and interface transparency rather than as a separate
leaderboard. The clinical targets are shared, but the input pathways are
not controlled equivalents: non-native-volume VLMs receive the
3D-evidence questions through 16-slice axial renderings for each
study-sequence, whereas M3D-LaMed and RadFM use their supported
native-volume pathways. The table therefore compares available model
interfaces on shared clinical targets, not the isolated value of one 3D
representation.
Open-ended Pass is shown only for configurations with released
factual-audit records; unavailable cells are marked with dashes rather than
inferred from closed-form scores.

\begin{table}[h]
\centering
\scriptsize
\setlength{\tabcolsep}{4pt}
\caption{Paper-reported 3D-evidence results provide compact numeric
provenance for the main 3D columns and the representative CI
visualization. Closed-form entries give exact accuracy (\%) with
bootstrap confidence intervals over the 1{,}123 closed-form items.
Open-ended entries report Pass \%; dashes indicate that released
factual-audit records are unavailable for that configuration.
Non-native-volume VLMs use 16-slice axial renderings; M3D-LaMed and
RadFM use their native-volume interfaces, so this table is an interface
comparison on shared clinical targets.}
\label{tab:threed-paper-reported}
\begin{tabular*}{\linewidth}{@{\extracolsep{\fill}}r l c r @{\hspace{1.0em}} r l c r@{}}
\toprule
\multicolumn{4}{c}{\textbf{Ranks 1--10}} &
\multicolumn{4}{c}{\textbf{Ranks 11--20}} \\
\cmidrule(lr){1-4}\cmidrule(lr){5-8}
Rank & Model & Closed [95\% CI] & Pass &
Rank & Model & Closed [95\% CI] & Pass \\
\midrule
1 & GPT-5.4 & 42.2 [39.6, 45.1] & 15.6 &
11 & Qwen2.5-VL-7B & 33.2 [30.5, 36.0] & 3.9 \\
2 & GPT-5 & 38.3 [35.4, 41.0] & 11.7 &
12 & Gemini 2.5 Pro & 32.0 [29.3, 34.7] & 13.0 \\
3 & GPT-5-mini & 37.5 [34.7, 40.3] & 10.4 &
13 & M3D-LaMed & 29.9 [27.0, 32.6] & 0.0 \\
4 & Qwen3-VL-4B & 35.4 [32.6, 37.9] & 5.2 &
14 & HuatuoGPT-Vision-7B & 29.1 [26.4, 31.7] & 0.0 \\
5 & Lingshu-7B & 35.4 [32.6, 38.1] & 3.9 &
15 & MedGemma-4B-it & 18.9 [16.7, 21.4] & 3.9 \\
6 & InternVL3-8B & 35.3 [32.3, 37.9] & -- &
16 & Janus-Pro-7B & 17.6 [15.5, 19.9] & 0.0 \\
7 & Qwen3-VL-30B & 34.8 [32.1, 37.5] & 11.7 &
17 & RadFM & 15.0 [13.0, 17.1] & 2.6 \\
8 & InternVL3-14B & 34.0 [31.4, 36.8] & -- &
18 & Qwen3.5-27B & 12.6 [10.5, 14.6] & 3.9 \\
9 & Qwen3-VL-8B & 33.4 [30.6, 36.1] & 9.1 &
19 & Claude Opus 4.6 & 1.2 [0.6, 1.9] & 6.5 \\
10 & InternVL3-78B & 33.3 [30.6, 36.3] & -- &
20 & LLaVA-Med-v1.5 & 0.0 [0.0, 0.0] & 1.3 \\
\bottomrule
\end{tabular*}
\end{table}

\subsection{Ranking sensitivity across evaluation axes}
\label{app:ranking-sensitivity}

\begin{figure}[h]
\centering
\includegraphics[width=0.88\linewidth]{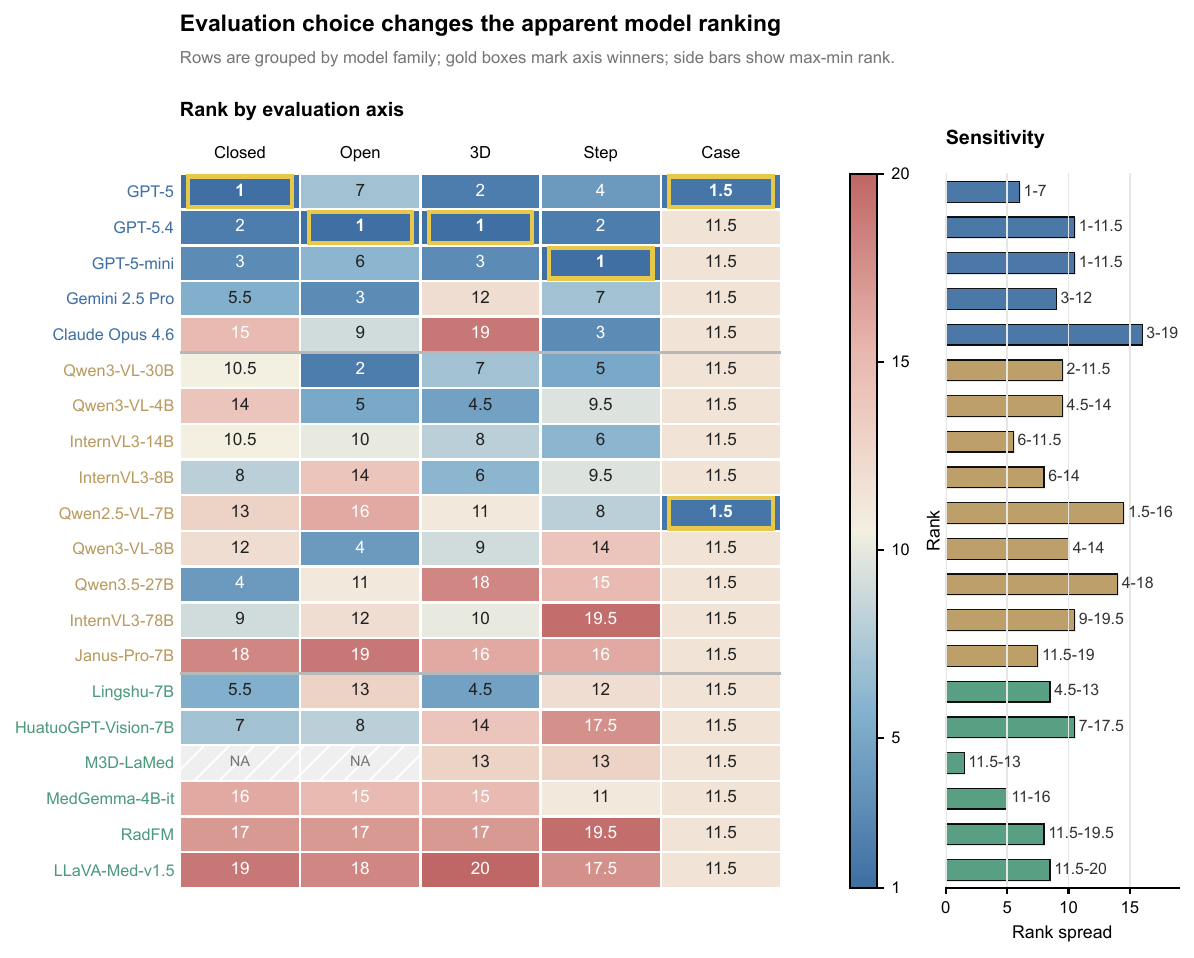}
\caption{Evaluation choice changes the apparent model ranking. Each
cell reports a model's rank on one evaluation axis, with lower ranks
better; the side bars summarize each model's rank spread across axes.
The figure is a sensitivity diagnostic, not a replacement for the
per-axis results.}
\label{fig:ranking-sensitivity}
\end{figure}

Figure~\ref{fig:ranking-sensitivity} summarizes how model order changes
across evaluation axes. Ranks are computed column-wise over eligible
main-paper rows, with missing entries excluded. The diagnostic point is
that BrainTRACE should not be reduced to a single aggregate score:
closed-form recognition, open-ended clinical acceptability, 3D-evidence
performance, Step Pass, and Case Success reward different aspects of
evidence use.

\FloatBarrier

\subsection{Calibrated level difficulty}
\label{app:calibrated-difficulty}

\begin{figure}[h]
\centering
\includegraphics[width=0.86\linewidth]{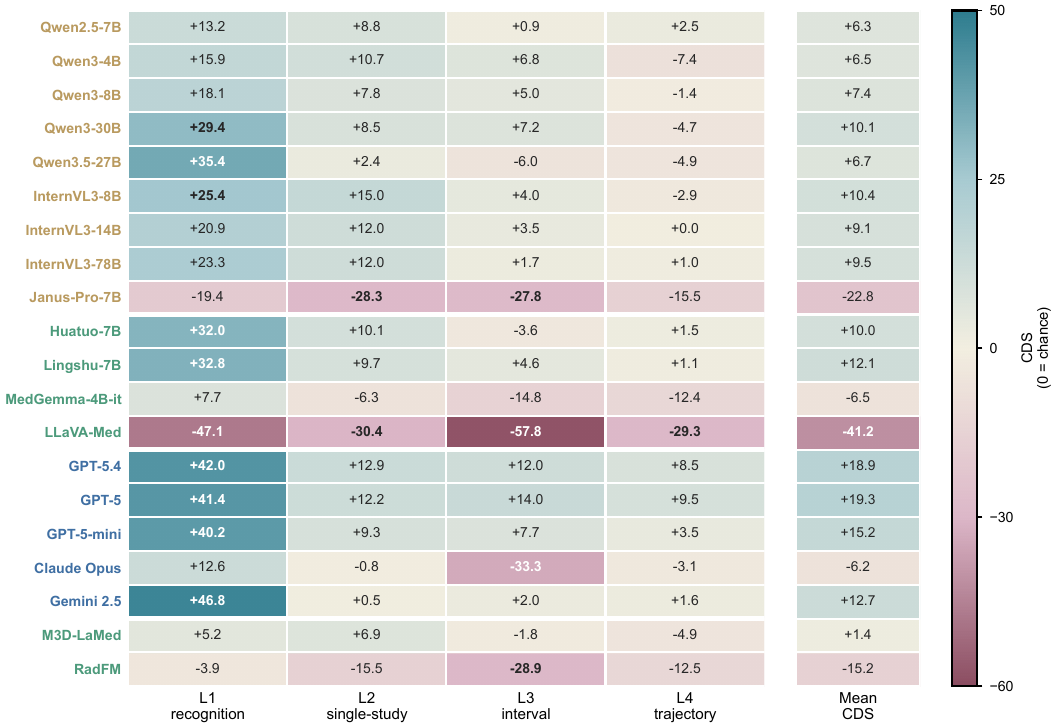}
\caption{Chance-calibrated difficulty shows the drop from recognition
to longitudinal reasoning after correcting for level-specific answer
floors. CDS sets each level's random baseline to 0 and perfect accuracy
to 100; plum cells are below chance, neutral cells are near chance, and
teal cells are above chance. Model names use the same family colors as
Figure~\ref{fig:ranking-sensitivity}.}
\label{fig:cds-levels}
\end{figure}

Figure~\ref{fig:cds-levels} reports a calibrated difficulty score
(CDS) for closed-form L1--L4 items. Raw accuracy is not directly
comparable across levels because the multiple-choice chance floor
differs by template mix. For model $m$ and level $\ell$, we define
\[
\mathrm{CDS}_{m,\ell}
=100 \times
\frac{\mathrm{Acc}_{m,\ell}-\mathrm{Chance}_{\ell}}
{100-\mathrm{Chance}_{\ell}},
\qquad
\mathrm{Chance}_{\ell}
=\frac{1}{|\mathcal{Q}_{\ell}|}
\sum_{q\in\mathcal{Q}_{\ell}}\frac{100}{|\mathcal{A}_q|},
\]
where $\mathcal{Q}_{\ell}$ is the set of closed-form questions at
level $\ell$ and $|\mathcal{A}_q|$ is the number of answer options for
question $q$. Thus 0 is the level-specific random baseline, 100 is
perfect accuracy, and negative values indicate below-chance behavior.

\FloatBarrier

The calibration prevents levels with higher answer-option floors from
appearing easier than they are. For example, GPT-5 has higher raw L3
accuracy than raw L2 (45.6\% vs.\ 32.9\%), but the corresponding CDS
values are close (+14.0 vs.\ +12.2) because L3 has a higher chance
floor. Conversely, Qwen3.5-27B has a seemingly moderate raw L3 accuracy
of 33.0\%, but its L3 CDS is -6.0 because this falls below the L3
random baseline; the same model remains clearly above chance on L1
(+35.4 CDS). The calibrated view therefore supports the main empirical
interpretation: recognition-level strength does not imply reliable
longitudinal reasoning.

%% file: sections/C_appendix_ablation.tex
\section{Evidence-channel ablation details}
\label{app:evidence-ablation}

This appendix clarifies the matched evidence-channel comparisons used in
Section~\ref{sec:evidence-ablation}. These comparisons are diagnostic:
they test whether accuracy on selected closed-form items depends on
longitudinal context, multi-sequence information, and visual evidence
when the question text, answer options, permitted non-image fields, and
scoring rule remain fixed. They are not intended as population-wide
performance estimates over all BrainTRACE items.

\subsection{Evidence retained in each comparison}
\label{app:ablation-conditions}

The latest-timepoint condition retains the current study but removes
prior longitudinal studies. Its paired accuracy drop measures dependence
on serial context, including interval change and trajectory evidence.
The T2-only condition retains the longitudinal T2 series but removes
non-T2 sequences. Its drop measures dependence on multi-sequence
contrast and sequence-specific findings, including enhancement or
susceptibility cues when those are needed by the reference. The no-image
condition withholds visual inputs while keeping the same question and
answer space. Its drop measures residual dependence on question text,
answer priors, or permitted non-image cues when rendered MRI evidence is
absent.

\subsection{Finding and interpretation}
\label{app:ablation-interpretation}

Figure~\ref{fig:bootstrap-ci-forest} gives the paired bootstrap
intervals for these evidence-channel drops. The intervals are computed
within each matched comparison, so the appendix uses them to support the
direction and stability of the drop rather than to introduce another
leaderboard. Across the retained comparisons, the paired drops remain
negative for both audited models. This supports the paper's main
interpretation that the selected BrainTRACE items require the retained
evidence channels, not only answer priors or generic question text.

The matched item sets were constructed to share a common full-evidence reference accuracy of 50.0\%, making the paired drops directly comparable within this diagnostic design. Because the ablation set is calibrated to this 50.0\% baseline, the results should be interpreted as paired diagnostic drops rather than population-level estimates. Abstentions in the no-image condition are counted as incorrect under the same closed-form exact-match rule used throughout the benchmark. The companion release provides the evaluated items for each comparison, ensuring that the denominators underlying the paired intervals remain auditable.

%% file: sections/D_appendix_cases.tex

\section{Qualitative failure cases}
\label{app:case-examples}

This appendix chapter is a compact case atlas for the factual-audit
protocol. It shows why fluent answers can fail when they reverse
laterality, miss or mislocalize a lesion, reverse interval direction, or
introduce an unsupported diagnostic category. The examples complement
the aggregate results by making the scoring rule inspectable.

The section presents paired GPT-5.4 success and failure examples across
the five clinical levels. Each pair uses the same task template and
illustrates a clinically material failure mode. Annotation legend:
\colorbox[HTML]{DBEAFE}{\color[HTML]{1E40AF}\small\textsf{blue}} = matched required facts between reference and response;
\colorbox[HTML]{DCFCE7}{\color[HTML]{166534}\small\textsf{green}} = correct clinical fact in reference;
\colorbox[HTML]{FDE2E2}{\color[HTML]{991B1B}\small\textsf{red}} = erroneous claim in model response.
Images are reproduced from the MR-RATE corpus under its CC BY-NC-SA 4.0 license.

Table~\ref{tab:case-failure-summary} summarizes the failure mode
illustrated at each clinical level.

\begin{table}[h]
\centering
\scriptsize
\setlength{\tabcolsep}{4pt}
\caption{The case atlas covers distinct clinically material failure
modes across the five task levels.}
\label{tab:case-failure-summary}
\begin{tabular}{@{}clllc@{}}
\toprule
Level & Task template & Failure mode & Error type & Images \\
\midrule
L1 & \texttt{one\_sentence\_image\_description} & Reversed laterality (left $\to$ right) & \texttt{wrong\_laterality} & 1 \\
L2 & \texttt{morphology\_and\_boundary} & Missed lesion / false negative & \texttt{missing\_finding} & 1 \\
L3 & \texttt{open\_interval\_change\_at\_site} & Decrease reported as increase & \texttt{wrong\_direction} & 6 \\
L4 & \texttt{open\_trajectory\_summary} & Mislocated + wrong trajectory & \texttt{wrong\_anatomy, wrong\_direction} & 3 \\
L5 & \texttt{full\_impression\_3to5\_sentences} & Wrong diagnostic category & \texttt{wrong\_diagnosis} & 6 \\
\bottomrule
\end{tabular}
\end{table}

\begin{figure}[h]
    \centering
    \includegraphics[width=\linewidth]{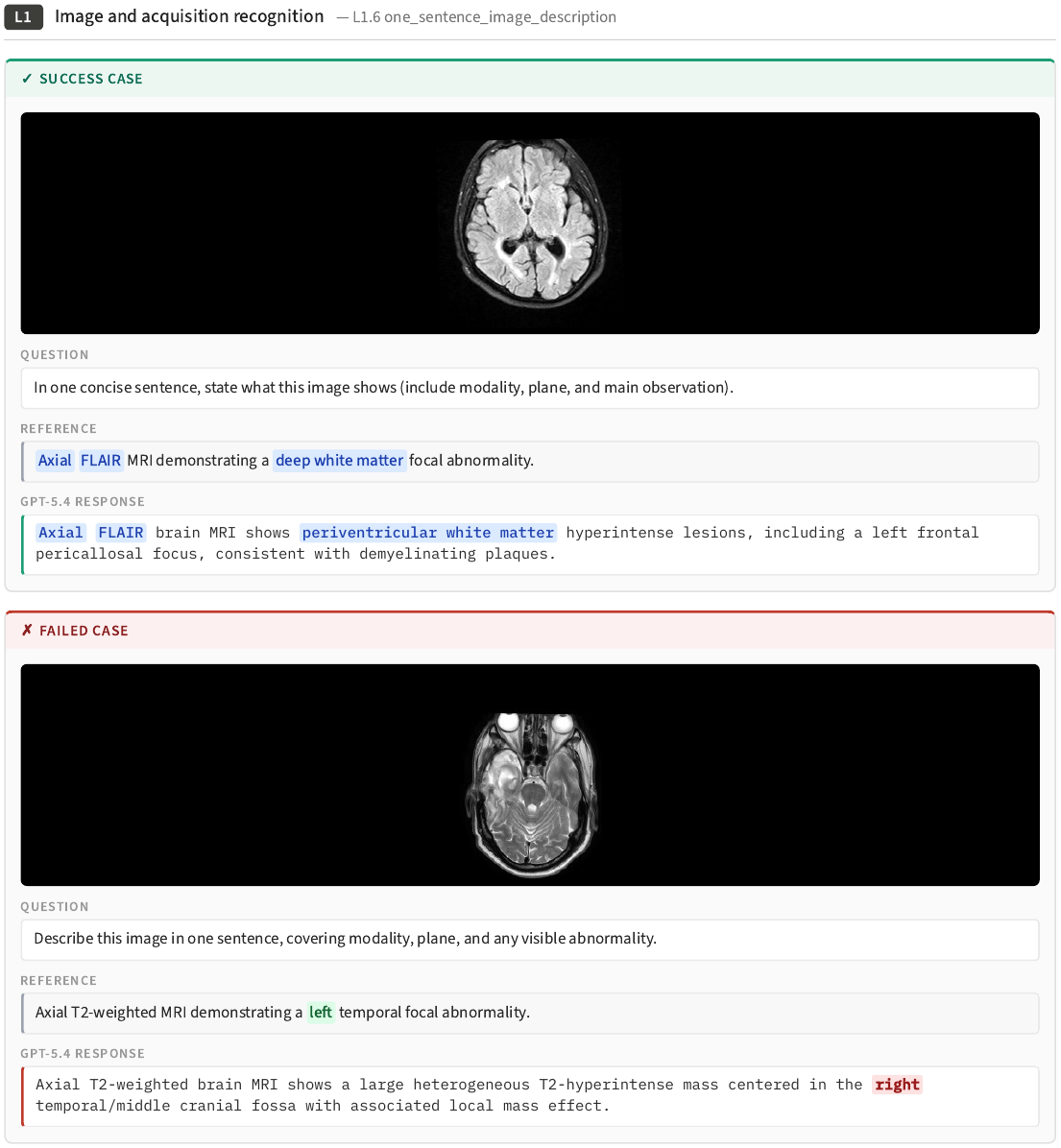}
    \caption{L1: Image and acquisition recognition (\texttt{L1.6 one\_sentence\_image\_description}). The model matches the required facts in the success case but reverses left/right laterality in the failure case. This is a clinically material contradiction under the factual-audit rule even when the rest of the description is fluent.}
    \label{fig:case-l1}
\end{figure}

\begin{figure}[h]
    \centering
    \includegraphics[width=\linewidth]{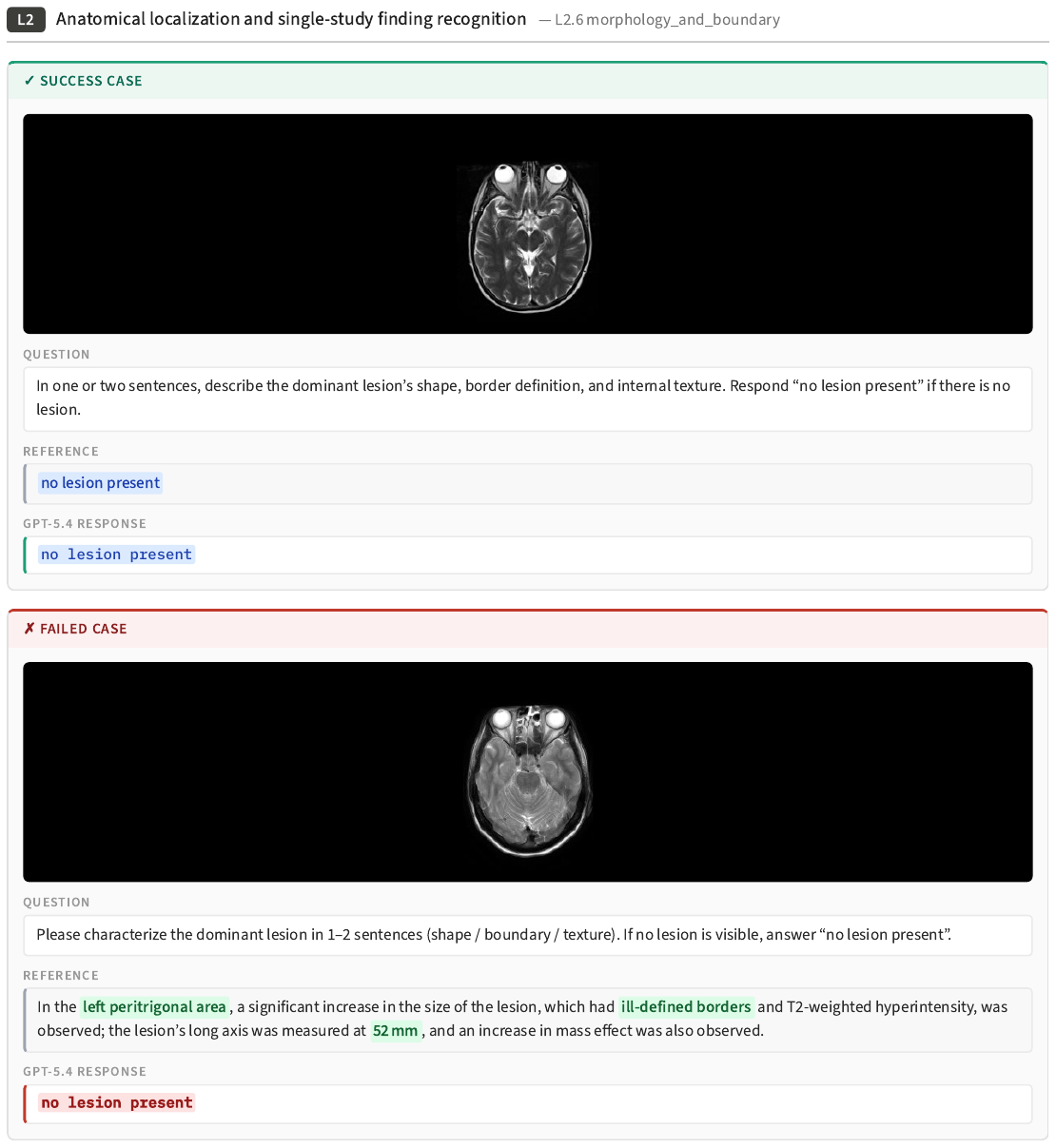}
    \caption{L2: Anatomical localization and single-study finding recognition (\texttt{L2.6 morphology\_and\_boundary}). Both cases use a single axial T2W input and receive the same model response. The success case is compatible with the reference, whereas the failure case omits a 52\,mm left peritrigonal lesion with ill-defined borders; the omission is a false negative rather than a wording difference.}
    \label{fig:case-l2}
\end{figure}

\begin{figure}[h]
    \centering
    \includegraphics[width=\linewidth]{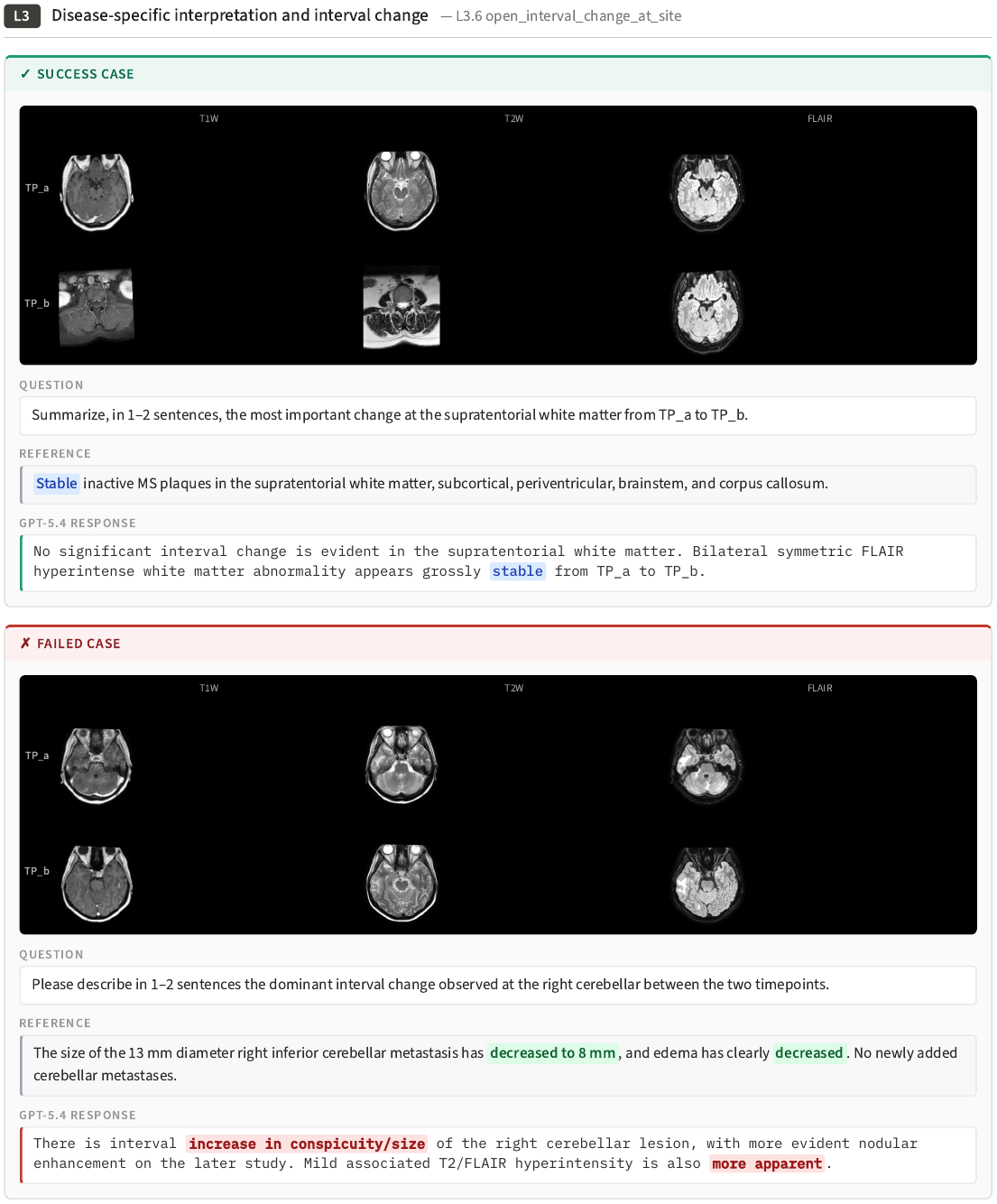}
    \caption{L3: Disease-specific interpretation and interval change (\texttt{L3.6 open\_interval\_change\_at\_site}). With longitudinal multi-sequence inputs (2 timepoints $\times$ 3 sequences), the model identifies stable lesions in the success case but reverses the interval direction in the failure case, reporting increased size when the metastasis decreased from 13\,mm to 8\,mm.}
    \label{fig:case-l3}
\end{figure}

\begin{figure}[h]
    \centering
    \includegraphics[width=\linewidth]{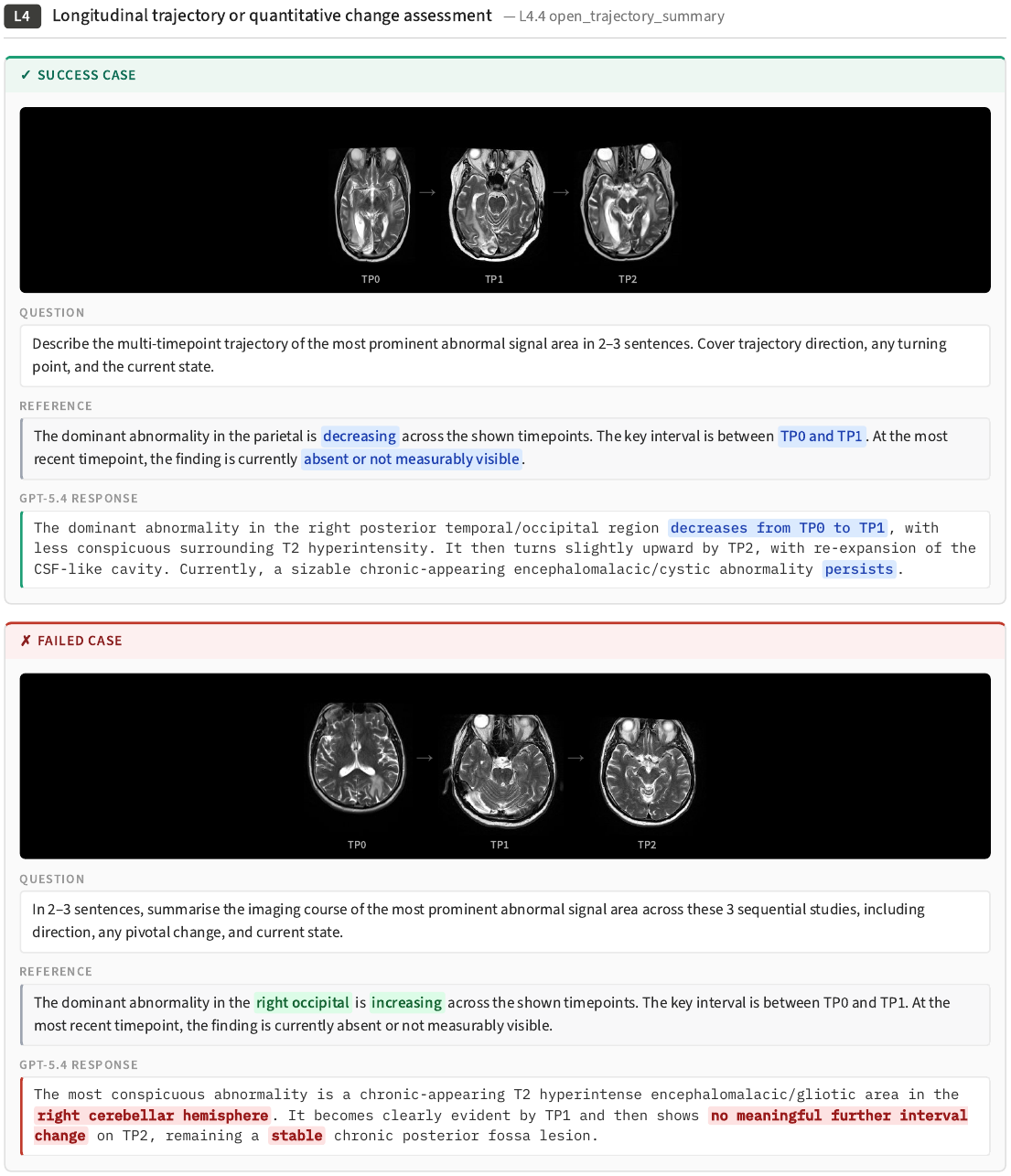}
    \caption{L4: Longitudinal trajectory or quantitative change assessment (\texttt{L4.4 open\_trajectory\_summary}). Across three serial T2W studies, the model traces a decreasing trajectory in the success case. In the failure case, it mislocalizes the finding to the cerebellar hemisphere rather than the right occipital region and misreads an increasing trajectory as stable.}
    \label{fig:case-l4}
\end{figure}

\begin{figure}[h]
    \centering
    \includegraphics[width=0.90\linewidth]{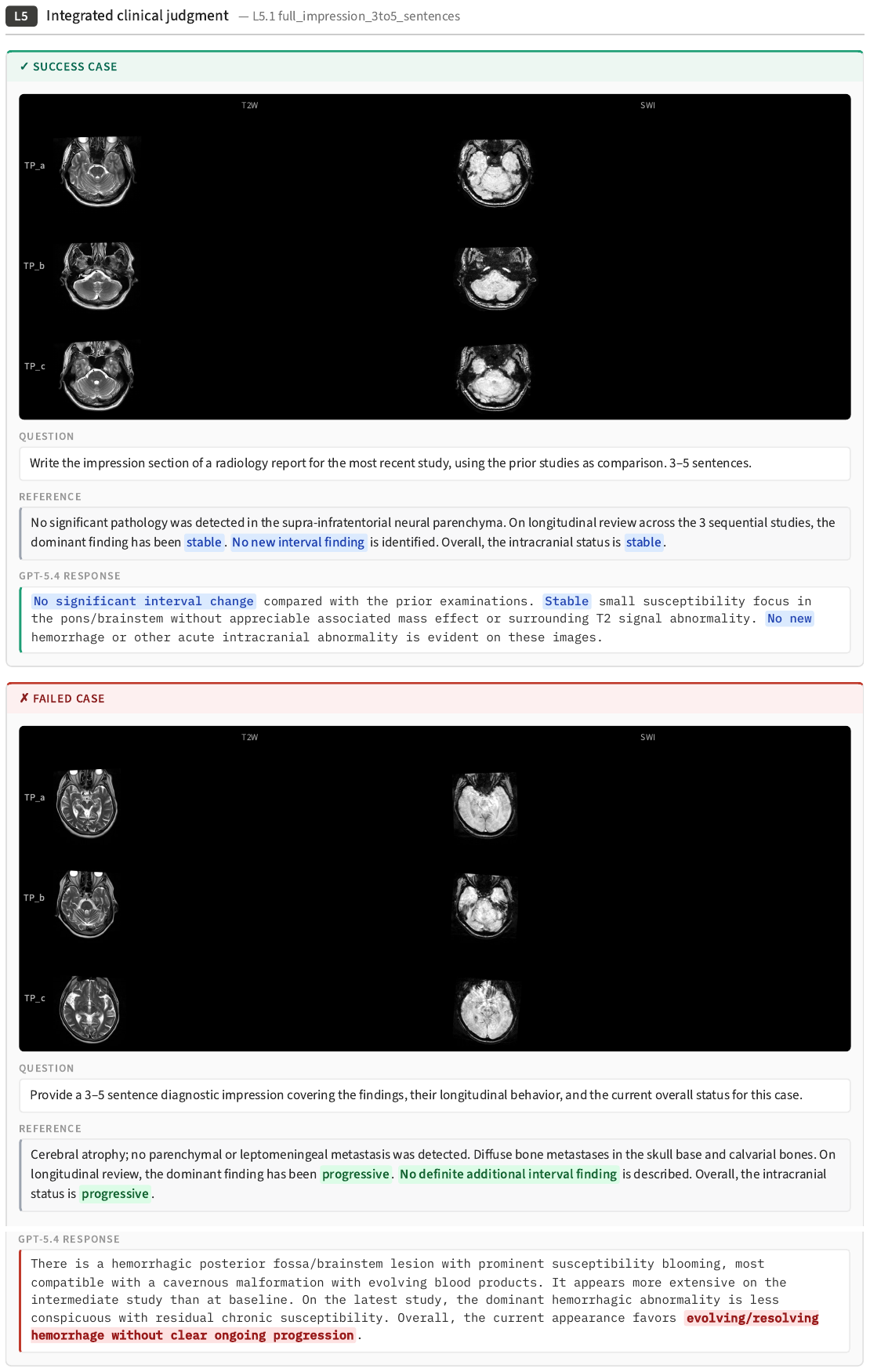}
    \caption{L5: Case-level and integrated clinical judgment (\texttt{L5.1 full\_impression\_3to5\_sentences}). Both cases present 3 timepoints $\times$ 2 sequences (T2W + SWI). The model identifies stable status with no new findings in the success case, but assigns the wrong diagnostic category in the failure case by interpreting progressive bone metastases as resolving hemorrhage.}
    \label{fig:case-l5}
\end{figure}

%% file: sections/05_limitations_ethics.tex
\section{Limitations and Responsible Use}
\label{app:limitations-responsible-use}
\label{sec:limitations-ethics}

BrainTRACE evaluates longitudinal reasoning over brain MRI follow-up
studies in an MR-RATE-derived cohort. Its task families assume
multi-sequence brain imaging, English language radiology reports, and
cases with sufficient follow-up, report, and segmentation-derived
evidence. It is not intended to represent other anatomies,
other languages, one-off clinical referrals with sparse longitudinal
context, or the full spectrum of neuroradiology indications. Its scores
are designed for reproducible
model comparison, not for estimating absolute clinical performance.
Users should report per-level, per-attribute, and acquisition subgroup
results rather than treating a single aggregate score as a clinical
capability measure.

The benchmark is report-grounded rather than independently re-read from
scratch. The references come from authentic MR-RATE clinical radiology
reports and segmentation-derived measurements, and released questions,
ground truth, and audit fields pass construction and consistency checks.
The case reasoning track further receives item-level review by an
ABR-certified radiation oncologist. Even with these controls, reports
are clinical documents rather than pixel-level annotations. A report may
compress timepoints, omit a visually apparent but clinically secondary
finding, or express uncertainty, and segmentation-derived measurements
can carry registration, acquisition, and algorithmic error. Scores
should be read as agreement with the reviewed, report-grounded
clinical references under the released benchmark specification, not as a
new adjudicated imaging diagnosis or deployment certification. The
primary open-ended scores use a GPT family factual audit judge, and a
small cross-family audit in Appendix~\ref{appx:judge-family-sensitivity}
shows judge family sensitivity for GPT-5.4 outputs. The
3D evidence track tests a clinically important evidence condition, but
its rendered 16-slice and native-volume pathways introduce an
input interface confound. These results should be read as
comparisons among available model interfaces on shared clinical targets,
not as a controlled ablation of 3D representation alone. BrainTRACE
releases evaluation records for audit, but it does not redistribute raw
imaging. Users must obtain access to the upstream MR-RATE
corpus~\citep{mrrate2026} and follow the corresponding data-use terms.